\documentclass{article} %
\usepackage{iclr2027_conference,times}

\usepackage{amsmath,amsfonts,bm}

\def\eqref#1{equation~\ref{#1}}

\def\1{\bm{1}}

\DeclareMathAlphabet{\mathsfit}{\encodingdefault}{\sfdefault}{m}{sl}
\SetMathAlphabet{\mathsfit}{bold}{\encodingdefault}{\sfdefault}{bx}{n}

\newcommand{\KL}{D_{\mathrm{KL}}}

\usepackage{hyperref}
\usepackage{url}

\usepackage{amssymb}
\usepackage{amsmath}
\usepackage{mathtools}
\usepackage{graphicx}
\usepackage{subcaption}

\title{In-context learning before specialisation: theory of learning in a solvable topic model}

\author{%
Francesco Cagnetta\thanks{Correspondence to: \texttt{fr.cagnetta@gmail.com}.}\\
Simplex
\And
Casper L. Christensen\\
Simplex, Astera Institute
}

\iclrfinalcopy
\begin{document}

\maketitle
\lhead{}
\renewcommand{\headrulewidth}{0pt}

\begin{abstract}
Sequence models can generalise in fundamentally distinct ways: extrapolating structures seen in training, or using their prompt to adapt to novel tasks.
The prevailing strategy determines how the model behaves beyond its training distribution, controlling potential capabilities and risks.
Recent work has characterised many different strategies theoretically, and empirically described the transition between them as the quality and quantity of pretraining data vary.
However, there is no learning theory for this competition, hence no fundamental understanding of \textit{why} one solution might be favoured over another.
Here we provide such a theory in a solvable topic model---documents drawn from a finite mixture of latent topics, mirroring the heterogeneity of modern pretraining data.
The dataset comes with two distinct solutions: a specialised predictor, that stores the finite topic mixture, and an in-context one, optimal for topics never seen during training.
Our theory predicts their emergence at distinct sample complexities: the in-context predictor is available early, while the specialised one emerges only after the training tokens outnumber the parameters of the data-generating process.
Trained transformers reproduce both scales and the separating crossover in remarkable agreement with the theory, with no fitted parameters.
\end{abstract}

\begin{figure}[h!]
  \centering
  \begin{subfigure}[t]{0.47\textwidth}
    \vspace{0pt}
    \centering
    \includegraphics[width=\linewidth]{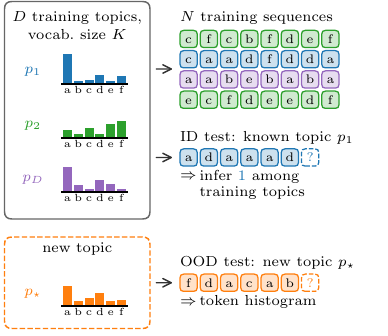}
    \label{fig:main-schematics}
  \end{subfigure}%
  \begin{subfigure}[t]{0.53\textwidth}
    \vspace{0pt}
    \centering
    \includegraphics[width=\linewidth]{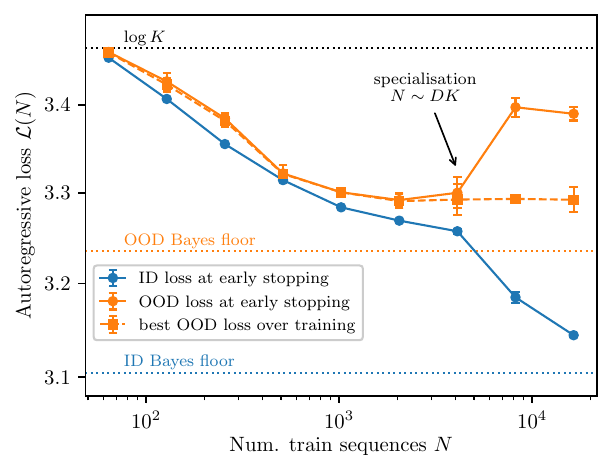}
    \label{fig:main-learningcurve}
  \end{subfigure}%
  \vspace{-10pt}
  \caption{
   \textbf{In-context learning and specialisation in a random topic model.}
   \textbf{Left:} A random topic model consists of $D$ topics, i.e. unigram probabilities $\mathbf{p}_z\in \Delta^{K-1}$ over a vocabulary of size $K$.
   Training sequences are obtained by sampling a topic $z$ uniformly, then $T$ tokens i.i.d. from $\mathbf{p}_z$.
   \textbf{Right:} Learning curve of a one-layer transformer trained for next-token prediction on a random topic model with $(K,D,T)\,{=}\,(32,512,128)$.
   Initially, both In-Distribution (ID, on the $D$ training topics, blue) and Out-Of-Distribution (OOD, on a new topic $\mathbf{p}_\star$, orange) descend with $N$ towards the entropy of the Bayes-optimal OOD predictor (orange dotted):
   the add-one estimator $r_t(a\mid X_{\leq t}) \,{=}\,(\#{\left\lbrace i\leq t \mid X_{i}=a\right\rbrace} + 1)/(t+K)$.
   Later, the predictor specialises to the training topics: the ID loss converges to the ID Bayes-optimal floor (blue dotted), while the OOD loss rises as in overfitting.
    Code: \url{https://github.com/fracagnetta/random-topic-model}.
  }
  \label{fig:main}
\end{figure}

\section{Introduction}\label{sec:intro}

Large language models generalise in ways that were not anticipated in training.
In particular, novel behaviours can be elicited simply via prompting, including few-shot learning of supervised tasks~\citep{brown2020language} or adopting different roles depending on a set of instructions~\citep{shanahan2023role}.
This In-Context Learning (ICL) mechanism is not captured by the classical learning-theoretic paradigm, where a model is evaluated on fresh samples from same the distribution it was trained on.
Nevertheless, ICL is a critical source of both generality and risk~\citep{anwar2024foundational} in modern AI systems, calling for a theory of what trained models do beyond their training distribution.

An early theoretical attempt framed ICL as implicit Bayesian inference of the latent concept behind the prompt~\citep{xie2022bayesian}.
This framework has found extensive support:
trained transformers behave as Bayes-optimal estimators for many families of tasks~\citep{garg2022what}, and how such estimators are represented and acquired during training is understood in simple settings~\citep{nichani2024how,edelman2024the,lu2025asymptotic,zhang2025training,bondaschi2026from}.
However, other settings admit distinct classes of predictors, and models transition between them as the training conditions vary~\citep{akyurek2023what}.
\citet{raventos2023pretraining} identified task diversity as the key control parameter:
When trained at low diversity, the transformer can only memorise the training tasks, acquiring general ICL abilities only above a certain task-diversity threshold.

The role of task diversity has been explored in synthetic data, revealing that the competition between the two behaviours extends along training time~\citep{singh2023transient,reddy2024mechanistic,nguyen2025differential,park2025competition,gibson2026distinct}, culminating in a hierarchical Bayesian framework that fits the model behaviour to a weighted mixture of the memorising and generalising predictors~\citep{carroll2025dynamics, wurgaft2025incontext}.
This framework is phenomenological: still missing is a \emph{microscopic} theory that predicts the favoured solution from data properties alone, and how much data it takes for one solution to give way to the other.

\paragraph{Our Contribution}
Here we provide this theory in a synthetic dataset containing the minimal required ingredients: a task diversity parameter $D$, a vocabulary of size $K$, a document length $T$ and a corpus of $N$ training documents.
Task diversity enters as a mixture of $D$ latent topics from which documents are generated~(\autoref{fig:main}, left).
In this setting, we provide:
\begin{itemize}
  \item \textbf{A solvable model of strategy competition.} We characterise analytically two distinct Bayes-optimal predictors: a \emph{specialised} one, relying on knowledge of the $D$ topics in the data-generating process, and an \emph{in-context} one, equivalent to the add-one estimator from statistics under our data model (\autoref{sec:rtm}).
  \item \textbf{A first-principles learning-curve theory for specialisation.} We derive the specialisation learning curve in a Bayesian learning setting: the excess loss over the asymptotic floor is $DK/(2NT)$ at leading order---a function of the number of training tokens per parameter of the data-generating process (\autoref{sec:special-theory}).
  \item \textbf{A statistical mechanism for the in-context phase.} The in-context predictor is linear in the token counts, and a learner that estimates this map from data acquires it at a sample complexity set by its $K^2$ parameters and by the information that a context carries about the next token: both independent of $D$. Hence, comparison with the specialisation scale gives a task-diversity threshold above which (see~\autoref{fig:main}) \emph{the in-context solution emerges first, it remains stable in a window that widens with task diversity, then it gives way to the specialised solution} (\autoref{sec:icl-theory}).
  \item \textbf{Parameter-free agreement with trained transformers.} Across vocabularies, task diversities, sequence lengths and corpus sizes, transformer in-distribution losses collapse as functions of $NT/DK$, converging to the specialised solution at the predicted rate. Non-trivial ICL performance develops at the predicted, $D$-independent sample complexity, then vanishes again with specialisation (\autoref{sec:experiments}).
  \item \textbf{Separation from classical spectral methods.} Reconstructing the topics from low-order moments~\citep{hsu2012spectral,anandkumar2014tensor} requires resolving a Marchenko--Pastur bulk of eigenvalues, which takes $N\sim D^2K$ samples: a worse scaling with task diversity than the Bayes-optimal $NT\sim DK$ followed by trained transformers (\autoref{ssec:spectral}).
\end{itemize}

\subsection{Related work}\label{sec:related}

\paragraph{Mixtures of tasks as ICL testbeds.}
Finite task mixtures underlie most theoretical studies of ICL: linear regression~\citep{garg2022what,akyurek2023what,raventos2023pretraining,lu2025asymptotic,zhang2025training,bordelon2026theory,letey2026pretraintest}, classification~\citep{reddy2024mechanistic,nguyen2025differential}, and mixtures of Markov~\citep{edelman2024the,park2025competition,bondaschi2026from,gibson2026distinct} and hidden Markov~\citep{riechers2025next,ray2026nonergodic} models.
The ``Balls \& Urns'' setting of \citet{wurgaft2025incontext} is precisely the random topic model studied here, and their memorising and generalising predictors are our specialised and in-context ones.
Our work complements their phenomenological Bayesian framework: where they assume a scaling law and a low-Kolmogorov-complexity bias, we derive the sample complexities of the two predictors from first principles.
Thus, our approach does not require estimating Kolmogorov complexities and fitting parameters, but rather prescribes a simple picture where the functional simplicity of the ICL solution promotes its emergence as long as it leads to a noticeable loss gain with fewer data.

\paragraph{Representations, Belief-state geometry and task vectors.} Due to next-token pretraining, transformers linearly represent the belief state over latent structure~\citep{shai2024transformers}.
In particular, mixtures of independent components, or \emph{non-ergodic} sources, give rise to a telescopic geometry, which sparsifies along the context~\citep{ray2026nonergodic}.
The random topic model is the simplest non-ergodic source, and our results predict the sample dependence of the resulting belief geometry. 
Beyond controlled data structures, ICL has been linked to task vectors encoding the inferred task~\citep{hendel2023incontext}, whose geometry distinguishes dual modes of task inference corresponding to our two predictors~\citep{yan2026task}.
Overall, these approaches describe \emph{how} a trained network implements the chosen strategies; ours predicts \emph{when} the data favours each.

\paragraph{Learning theories of ICL.}
Most analytical studies of learning focus on the infinite-data and infinite-diversity limit.
\citet{lu2025asymptotic} is closest to us: they solve linear attention trained on Gaussian regression in a proportional asymptotic limit, obtaining sharp learning curves and a genuine transition with increasing task diversity.
However, since their linear model cannot fit the specialised predictors exactly, they did not observe the crossover from in-context to specialised predictor as the number of data increases.
Furthermore, our setup can be extended to mixtures of Markov chains~\citep{gibson2026distinct} and generic hidden Markov models~\citep{riechers2025next}.

\section{Data model and optimal predictors}\label{sec:rtm}

This section introduces our synthetic model of mixture data and the two reference predictors representing the \emph{specialised} and \emph{in-context solutions}.
The solutions are distinguished by their next-token-prediction loss profiles.
As shown in~\autoref{fig:id-vs-ood-profiles}, trained transformers behave like the specialised predictor at low data diversity, while converging to the in-context one at high diversity.
The rest of the paper explains why, and on which scale of the corpus size $N$ each solution is selected.

\paragraph{Random Topic Model}
A dataset is specified by a vocabulary size $K$, a number of topics $D$ and a document length $T$.
Each topic is a probability distribution over the vocabulary, $\mathbf{p}_z\in \Delta^{K-1}$ with entries $p_z(a)$ for $a\,{=}\,1,\dots,K$, drawn independently and uniformly from the simplex, $\mathbf{p}_z\sim\mathrm{Dir}_K(\mathbf{1})$ for $z\,{=}\,1,\dots,D$, and then held fixed.
We collect the topics as the columns of the topic matrix $P\,{=}\,[\mathbf{p}_1,\dots,\mathbf{p}_D]\in\mathbb{R}^{K\times D}$.
A document is generated by drawing a topic index $Z$ uniformly from $\lbrace 1,\dots,D\rbrace$, then $T$ tokens $X_1,\dots,X_T\in\lbrace 1,\dots,K\rbrace$ independently from $\mathbf{p}_Z$ (\autoref{fig:main}, left).
Given the topic, the likelihood of a context $X_{\leq t}\,{=}\,(X_1,\dots,X_t)$ depends only on its histogram, the count vector
\begin{align}
  C(X_{\leq t}) \in \mathbb{N}^{K},\qquad C_a(X_{\leq t})\,{:=}\,\#\lbrace s\,{\leq}\,t : X_s\,{=}\,a\rbrace.
\end{align}
The training corpus consists of $N$ documents, all generated with the same topic matrix $P$.

\paragraph{Data versus topic distribution.}
Two kinds of averages appear throughout.
$\mathbb{E}[\cdot]$ denotes the expectation over documents at fixed topic matrix $P$, i.e. over the topic index $Z$ and the tokens: this is the average that defines the loss of any predictor, including a trained transformer.
$\overline{\,\cdot\,}$ denotes the further average over the draw of the topics.
Unless stated otherwise, the curves in the figures are averaged over independent realisations of the topic matrix $P$.

\paragraph{Next-token prediction, In- and Out-Of-Distribution.}
We consider two test distributions.
In-distribution (ID) documents are generated as the training ones, from one of the $D$ topics of $P$.
Out-of-distribution (OOD) documents are generated from a fresh topic $\mathbf{p}_\star\sim\mathrm{Dir}_K(\mathbf{1})$, drawn from the same prior as the training topics but independently, and hence almost surely different from all of them.
A next-token predictor $\mathbf{f}$ maps each context $X_{\leq t}$ to a distribution $\mathbf{f}_t(X_{\leq t})\in\Delta^{K-1}$ over $X_{t\,{+}\,1}$. We measure ID performance by the next-token loss at context length $t$, or $t$-gram entropy
\begin{align}\label{eq:ntp-loss}
  \mathcal{L}_t[\mathbf{f}] := \mathbb{E}\left[ -\log f_t\!\left(X_{t+1}\mid X_{\leq t}\right) \right],
  \qquad t\,{=}\,0,\dots,T\,{-}\,1,
\end{align}
where the expectation is over fresh documents generated from the $D$ training topics.
The same expectation over a fresh topic $\mathbf{p}_\star$ defines the OOD loss $\mathcal{L}^\star_t[\mathbf{f}]$.
The usual autoregressive loss $\mathcal{L}_{\rm AR}$ is the average of $\mathcal{L}_t$ over positions.

\subsection{The specialised predictor}\label{ssec:specialised}

A learner that knows the topic matrix $P$ still does not observe the topic $Z$ of a given document, and must infer it from the context.
The Bayes-optimal next-token distribution is the posterior mixture of the topics,
\begin{align}\label{eq:specialised}
  \mathbf{q}_t(X_{\leq t}) := \sum_{z=1}^D \gamma_t(z)\,\mathbf{p}_z,\qquad \gamma_t(z) := \mathbb{P}\left( Z\,{=}\,z \mid X_{\leq t}\right)
  = \frac{\prod_{a=1}^K p_z(a)^{C_a}}{\sum_{y=1}^D \prod_{a=1}^K p_y(a)^{C_a}},
\end{align}
where the posterior over the topic $\gamma_t(z)$ depends on the context only through the counts.
$\mathbf{q}_t$ is the true conditional distribution of $X_{t+1}$ given the context, hence it minimises $\mathcal{L}_t$ over all predictors (\autoref{app:specialised}).
We call it \emph{specialised}, as it is built from the $D$ topics realised in training.

The cross-entropy loss splits into an entropy plus a Kullback--Leibler divergence,
\begin{align}\label{eq:specialised-loss}
  \mathcal{L}_t[\mathbf{q}] = \frac{1}{D}\sum_{z=1}^D H(\mathbf{p}_z) + \mathbb{E}\left[\KL\left(\mathbf{p}_Z \,\Vert\, \mathbf{q}_t\right)\right],
\end{align}
both averaged over the unknown topic.
The first term is the loss of an oracle that observes the topic, and no predictor can go below it: it is the \emph{entropy rate} of the data-generating process.
For sufficiently large $D$, the entropy rate converges to the average over draws of the topic matrix $\overline{H(\mathbf{p})}$, which is $\log K - 1 + \gamma_{\mathrm{E}} + O(1/K)$ for large $K$ (\autoref{app:topic-stats}), with $\gamma_{\mathrm{E}}\simeq 0.57721$ Euler's constant.
The second term is the excess loss due to the uncertainty on the topic and, by~\autoref{eq:specialised}, it is controlled by the posterior mass on the wrong topics $y\,{\neq}\,z$.
As shown in~\autoref{app:specialised}, the mass on a wrong topic $y$ decays exponentially in $t$, at a rate given by the divergence $\KL(\mathbf{p}_z\Vert\mathbf{p}_y)$.
For random topics this divergence is of order one (\autoref{app:topic-stats}), so that discriminating the true topic from the other $D\,{-}\,1$ requires a context of length $t\sim\log D$ (\autoref{app:specialised}).
This scale is visible in the left panel of~\autoref{fig:id-vs-ood-profiles}.

\subsection{The in-context predictor}\label{ssec:in-context}

Consider now an OOD document, generated from a new topic $\mathbf{p}_\star$ that does not belong to the topic matrix $P$.
A learner that knows the prior $\mathrm{Dir}_K(\mathbf{1})$ but not $\mathbf{p}_\star$ must infer from the context the topic itself, rather than its index within a known dictionary.
By Dirichlet categorical conjugacy, the posterior over $\mathbf{p}_\star$ given $X_{\leq t}$ is $\mathrm{Dir}_K(C_1{+}1,\dots,C_K{+}1)$, and the Bayes-optimal next-token distribution is its mean, the add-one smoothing of the histogram (\autoref{app:in-context})
\begin{align}\label{eq:in-context}
  r_t\!\left(a\mid X_{\leq t}\right) := \mathbb{P}\left( X_{t+1}\,{=}\,a \mid X_{\leq t}\right) = \frac{C_a(X_{\leq t}) + 1}{t + K}.
\end{align}
We call it \emph{in-context} as it does not depend on the topic matrix $P$, and thus applies unchanged to documents from seen or unseen topics.

\begin{figure}[h!]
  \centering
  \includegraphics[width=\linewidth]{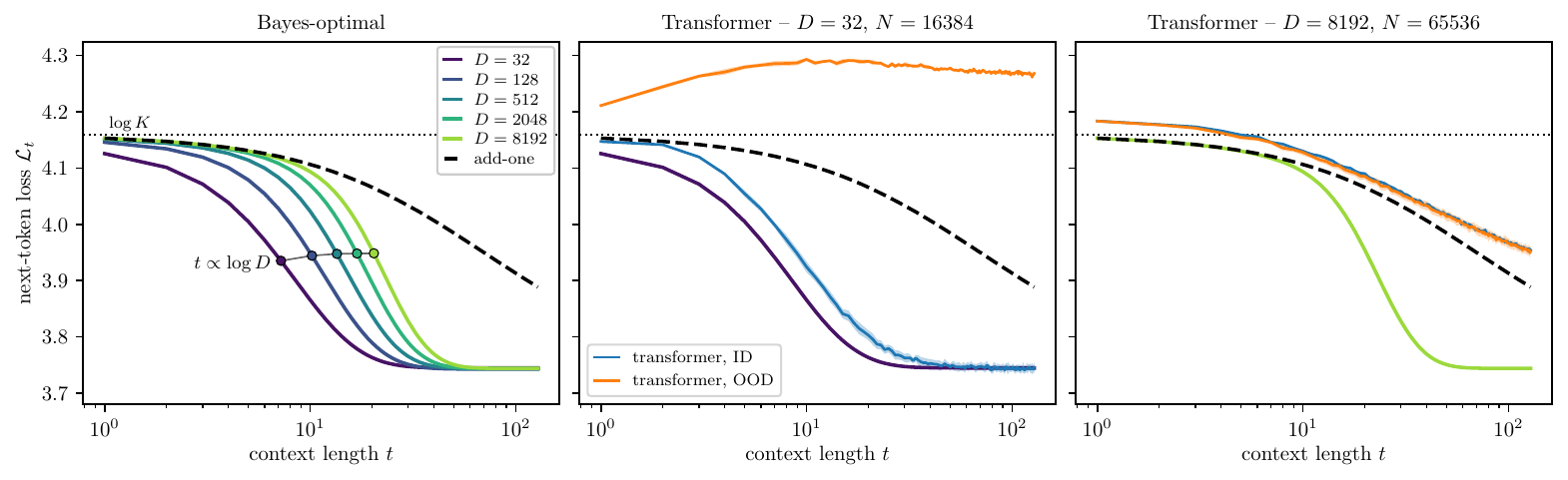}
  \vspace{-10pt}
  \caption{
   \textbf{Next-token loss profiles of the two Bayes-optimal predictors and of trained transformers}, for $K\,{=}\,64$ and $T\,{=}\,128$.
   \textbf{Left:} ID loss $\overline{\mathcal{L}_t[\mathbf{q}]}$ of the specialised predictor for $D$ from $32$ to $8192$ (colour), averaged over realisations of the topic matrix, and OOD loss $\mathcal{L}^\star_t[\mathbf{r}]$ of the in-context predictor (black dashed), which does not depend on $D$.
   Dotted line: $\log K$.
   Markers: context length at which each specialised profile has covered half of its descent, growing as $\log D$ (grey guide; \autoref{ssec:specialised}).
   \textbf{Middle and right:} the same two references at $D\,{=}\,32$ and $D\,{=}\,8192$, with the ID (blue) and OOD (orange) loss profiles of a one-layer transformer trained on $N$ documents at that $D$, mean $\pm$ s.e. over four runs with independent topic matrix, corpus and initialisation, at the checkpoint of minimal validation loss.
   At $D\,{=}\,32$ the transformer matches the specialised predictor in distribution and stays above $\log K$ out of distribution; at $D\,{=}\,8192$ both profiles follow the in-context predictor.
  }
  \label{fig:id-vs-ood-profiles}
\end{figure}

The OOD loss of $\mathbf{r}_t$ decomposes as in~\autoref{eq:specialised-loss}, with the unseen topic in place of $\mathbf{p}_Z$,
\begin{align}\label{eq:in-context-loss}
  \mathcal{L}^\star_t[\mathbf{r}] = \overline{H(\mathbf{p})} + \mathbb{E}\left[\KL\left(\mathbf{p}_\star\,\Vert\,\mathbf{r}_t\right)\right]
  = \overline{H(\mathbf{p})} + \frac{K-1}{2t} + O\!\left(\frac{\log t}{t^2}\right),
\end{align}
where the expansion holds at fixed $K$ for $t\gg K$ (\autoref{app:in-context}).
Simply put, the histogram needs to include at least order $K$ tokens to depart from a trivial prediction $1/K$.
Note that, since $\mathbf{p}_\star$ is drawn from the same prior as the topics of $P$, the entropy floor is the same for the two predictors, while the approach is significantly different (see~\autoref{fig:id-vs-ood-profiles}, left).
Identifying one of $D$ known topics requires $\log D$ nats of evidence, which the context delivers at a constant rate per token, hence the exponential convergence of \autoref{ssec:specialised}.
Estimating a new topic is instead a $(K{-}1)$-dimensional problem, and the excess in~\autoref{eq:in-context-loss} is the classical half of the number of parameters per observation behaviour.

\section{Learning-curve theory of specialisation}\label{sec:special-theory}

The training corpus contains only the $D$ topics in the matrix $P$.
Hence, a sufficiently expressive, well-optimised machine learning model \emph{must} converge to the specialised solution~\autoref{eq:specialised} in the infinite-data and training time limit.
But \emph{how many training documents does it take to specialise?}

\paragraph{Bayesian learning}
We answer this question with a Bayesian learner that uses the training corpus optimally:
It knows the model class, including $D$ and $K$, but not the realised topic matrix $P$, and observes a collection $\mathcal{C}_N$ of $N$ length-$T$ training documents, together with the test context.
The training corpus informs a posterior $\pi(dP|\mathcal{C}_N)$ over the topic matrix $P$, proportional, by Bayes' rule, to prior $\mu(dP)$ times the joint probability of the corpus under $P$, $\mathbb{P}\left\lbrace\mathcal{C}_N\right\rbrace$.
The Bayes-optimal learner is the posterior mean of the $P$-dependent specialised predictor from~\autoref{eq:specialised},
\begin{align}\label{eq:bayes-learner}
  \hat{\mathbf{q}}_t(X_{\leq t}) \propto \int \mathbf{q}_t(X_{\leq t}; P) \, \pi(dP|\mathcal{C}_N, X_{\leq t}),
\end{align}
where the posterior also includes the test context $X_{\leq t}$ (see~\autoref{app:bayes-learner} for details).
Note that, after averaging over the draw of the topics, no predictor built from the same data has a lower loss than $\hat{\mathbf{q}}$.
Its loss is the statistical limit of specialisation from $N$ training documents, hence a natural reference against which a transformer trained on the same corpus can be measured.

\paragraph{Learning curve and mutual information.}
The learning curve of specialisation is the excess loss of the Bayes-optimal learner over the specialised predictor, averaged over topic matrices,
\begin{align}\label{eq:learning-curve-def}
  \varepsilon_t(N) := \overline{\mathcal{L}_t[\hat{\mathbf{q}}] - \mathcal{L}_t[\mathbf{q}]}\,,
\end{align}
Averaging over the topics gives it an information-theoretic meaning: $\varepsilon_t(N)$ is the conditional mutual information between $P$ and the next token, given corpus and context (\autoref{app:bayes-learner}),
\begin{align}\label{eq:learning-curve-information}
  \varepsilon_t(N) = I(P;X_{t+1}\mid X_{\leq t}, \mathcal{C}_{N}) \Rightarrow \sum_{t=0}^{T-1}\varepsilon_t(N) = I(P;\mathcal{C}_{N+1}) - I(P;\mathcal{C}_N),
\end{align}
where $\mathcal{C}_{N+1}$ adds the test document to the corpus.
The excess loss of the learner on the test document equals the information that this document adds about the topic matrix.
Thus, the learning curve quantifies how fast the corpus reveals $P$.

\paragraph{The $DK/NT$ law.}
The topic matrix has $D(K{-}1)$ free parameters, and a corpus of $N$ documents reveals them at the rate of parametric statistics: under identifiability and regularity, $I(P;\mathcal{C}_N) = (D(K-1)/2)\log N + O(1)$~\citep{clarke1990information,clarke1994jeffreys}.
The increment gives the leading term of the learning curve at fixed $D$ and $K$ (\autoref{app:bayes-learner}),
\begin{align}\label{eq:dk-nt-law}
  \frac{1}{T}\sum_{t=0}^{T-1}\varepsilon_t(N) = \frac{D(K-1)}{2NT} + o\!\left(N^{-1}\right).
\end{align}
Specialisation thus requires $NT\gg DK$, and the learning curve is a function of the number of training tokens per parameter, $x\,{:=}\,NT/(DK)$, decaying as $x^{-1}$.
As shown in~\autoref{sec:experiments}, trained transformers specialise on the same sample-complexity scale, and with the same decay.
Two remarks.
First, identifiability, i.e. that the law of a document determines $P$ up to a relabelling of the topics, requires a sufficiently long document and a \emph{generic} topic matrix.
For $D\leq K$, $T\,{\geq}\,3$ suffices: the columns of $P$ are linearly independent almost surely under the topics draw, and the third-order moment of the tokens has a unique decomposition with the topics as factors~\citep{kruskal1977threeway,anandkumar2014tensor}.
For larger $D$ the same holds for the lifted topics $\mathbf{p}_z^{\otimes m}$, which are generic in the same sense, i.e. linearly independent, as long as $D\leq\binom{K+m-1}{m}$, the dimension of the space of symmetric tensors, at the price of longer documents, $T\,{\geq}\,2m{+}1$ (\autoref{app:bayes-learner}).
For instance, pairs of tokens cover $K<D\leq K(K{+}1)/2$ with $T\,{\geq}\,5$.
Secondly, the learning curve is an average over topic matrices, whereas an experiment runs on individual realisations.
Comparing the two assumes that the excess loss of a given realisation is close to its average, i.e. typicality of the random topic matrix.
An empirical check of this typicality is the spread across independent realisations in \autoref{sec:experiments}.

\paragraph{The same law at fixed, large position.}
Equation~\ref{eq:dk-nt-law} concerns the loss averaged over positions; at a fixed position $t$ the same law follows from a simple picture.
Beyond the resolution scale $\log D$ of \autoref{ssec:specialised}, the learner can identify the topic of the test document, and likewise the topics of the training documents if $T\gg\log D$.
It is then left with estimating the token distribution of the test topic from the training documents of that topic, about $N/D$ of them, i.e. from $m\approx NT/D$ tokens: this is the problem solved by the in-context predictor of \autoref{ssec:in-context}, with $m$ in place of $t$, whose excess $(K-1)/(2m)$ becomes $D(K-1)/(2NT)$.
The $t$-gram loss at large $t$ therefore obeys the same law as the average, and \autoref{app:bayes-learner} makes this rigorous at fixed $P$, under conditions on $t$, $T$ and $N$; \autoref{fig:tgram-scaling} compares it with the $T$-gram learning curves of trained transformers.

\section{When does ICL emerge?}\label{sec:icl-theory}

\paragraph{Simplicity of the in-context predictor.}
If any learner eventually converges to the specialised predictor, \emph{when does ICL emerge?}
Both predictors of \autoref{sec:rtm} are functions of the counts, but of very different complexity.
The in-context predictor is linear up to normalisation,
\begin{align}\label{eq:add-one-linear}
  \mathbf{r}_t(X_{\leq t}) \propto \mathbf{C}(X_{\leq t}) + \mathbf{1},
\end{align}
whereas the specialised predictor is exponential in the counts (\autoref{eq:specialised}).
Then, simplicity bias suggests that the in-context predictor is easier to learn.
$\mathbf{r}_t$ represents a point in the family of $K^2$ linear maps from counts to next-token probability, whereas random topic models $P$ lie in $D(K\,{-}\,1)$ dimensions.
The estimation error follows from the same argument as \autoref{sec:special-theory}, in the form appropriate to a fitted parametric family rather than a Bayesian posterior (\autoref{app:icl}).
Simply put, a learner estimating the $K^2$ parameters of the linear map from a corpus of $NT$ tokens pays an excess $K^2/(2NT)$ over the optimal in-context solution, again proportional to the number of parameters per training token.

\paragraph{Estimation penalty versus loss gain.}
This excess must be compared with the gain of the add-one rule over the uniform predictor, equal to the mean mutual information between the next token and its context (\autoref{app:icl}),
\begin{align}\label{eq:icl-gain}
  G(K,T) \coloneq \log K-\frac1T\sum_{t<T}\mathcal{L}^\star_t[\mathbf{r}] = \frac{1}{T}\sum_{t=0}^{T-1} I\left(X_{t+1}; X_{\leq t}\right).
\end{align}
The new solution can only appear once the error due to finite data $K^2/(2NT)$ is smaller than the gain, i.e. for
\begin{align}\label{eq:n-icl}
  N \gtrsim N_{\mathrm{icl}} := \frac{K^2}{2T\,G(K,T)},
\end{align}
independently of the task diversity $D$.

\paragraph{Task-diversity threshold.}
By \autoref{eq:dk-nt-law}, the specialised solution emerges at $N_{\mathrm{spec}}\,{=}\,DK/T$.
Hence, if
\begin{align}\label{eq:diversity-threshold}
  D \gg \frac{K}{2\,G(K,T)},
\end{align}
then the in-context solution emerges before the specialised one, and persists in the window $(N_{\mathrm{icl}}, N_{\mathrm{spec}})$.
The resulting picture is that of \autoref{fig:main}, right.
In the $(N_{\mathrm{icl}}, N_{\mathrm{spec}})$ window the learner applies the same rule to seen and unseen topics alike, thus OOD and ID loss decay together towards the OOD optimum $\mathcal{L}^\star_t[\mathbf{r}]$.
At $N_{\mathrm{spec}}$, the learner resolves the dictionary: the ID loss descends to the entropy rate, and the OOD loss rises towards $\mathcal{L}^\star_t[\mathbf{q}]$---OOD loss of the specialised predictor.

\section{Transformer experiments}\label{sec:experiments}

This section collects empirical learning curves of one-block transformers trained with Adam on data generated from random topic models of varying $K$ and $D$.
Experiments with varying $T$ are reported in~\autoref{app:varying-T}.
Since we are testing the scaling of performance with the number of training tokens, we conducted extensive hyperparameter search, looking for the best configuration of model size, learning rate, batch size and training time at fixed depth and optimiser.
Further details of architectures and training algorithms are reported in~\autoref{app:experiments}.

\begin{figure}[h!]
  \centering
  \includegraphics[width=\linewidth]{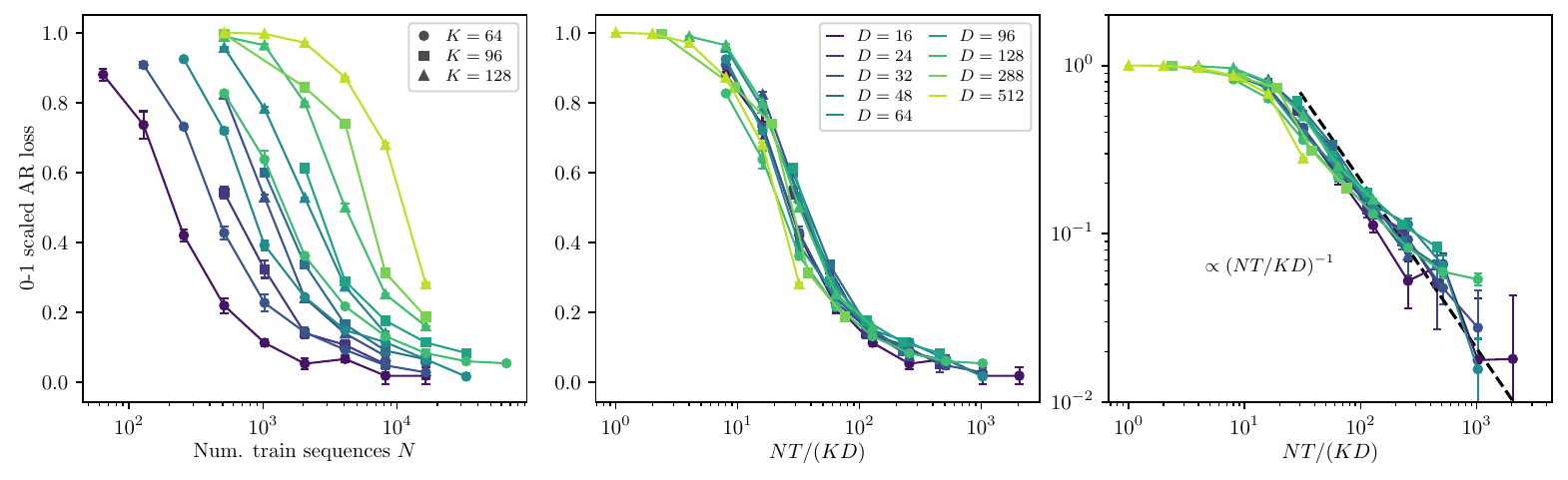}
  \vspace{-10pt}
  \caption{
   \textbf{Transformers match the Bayes-optimal specialisation learning curve.}
   \textbf{Left:} Scaled ID autoregressive loss $(\mathcal{L}_{\rm AR}(N)-\mathcal{L}_{\rm AR}(\infty))/(\log{K}-\mathcal{L}_{\rm AR}(\infty))$ versus number of training sequences for one-block transformers trained on random topic model data with varying $(K,D)$, $T\,{=}\,128$.
   The scaled loss is $1$ at the uniform predictor $1/K$, $0$ at the specialised predictor.
   \textbf{Middle:} Rescaling the $x$-axis as the number of tokens per parameter of the data-generating process collapses all the curves onto a single functional form.
   \textbf{Right:} The decay towards $0$ (black dashed) shows remarkable agreement with the $x^{-1}$ rate of the Bayes-optimal learning curve.
  }
  \label{fig:learning-curve-scaling}
\end{figure}

\autoref{fig:learning-curve-scaling} shows that transformers learn random topic models at the same rate as the Bayesian learner of~\autoref{sec:special-theory}.

\begin{figure}[h!]
  \centering
  \includegraphics[width=\linewidth]{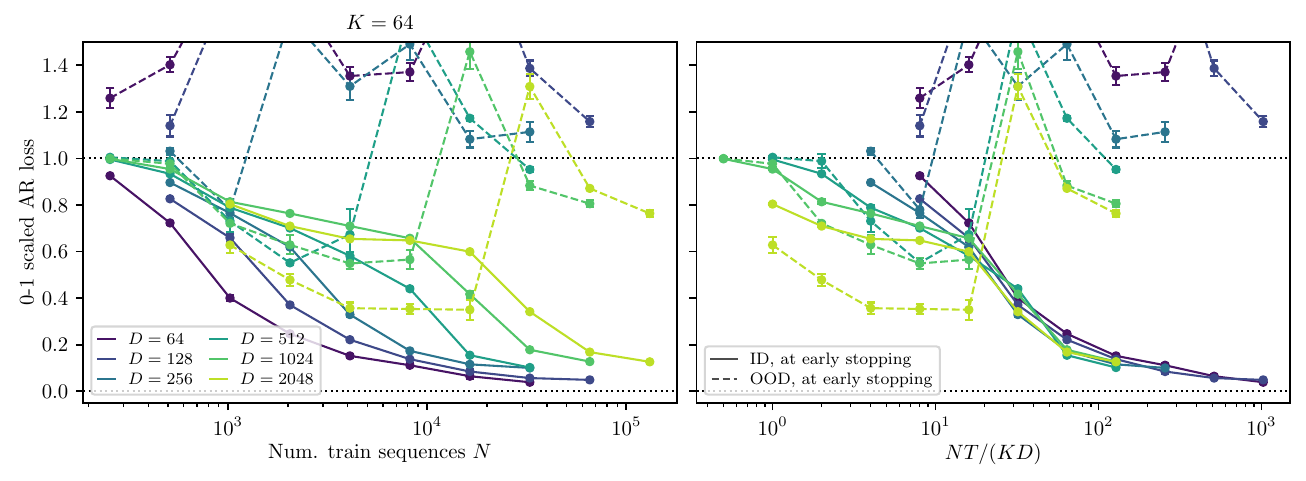}
  \vspace{-10pt}
  \caption{
   \textbf{In-context learning gives way to specialisation at the parametric scale $NT\sim DK$.}
   \textbf{Left:} Scaled ID (solid) and OOD (dashed) autoregressive losses, both subtracting the ID loss floor for comparison.
   \textbf{Right:} Rescaling the $x$-axis as the number of tokens per parameter of the data-generating process collapses both the descent of ID losses and the ascent of OOD losses.
  }
  \label{fig:ood-window}
\end{figure}

Even above the task-diversity threshold~\autoref{eq:diversity-threshold}, specialisation kicks in as soon as the number of tokens $NT$ surpasses the number of parameters $DK$.
This is clearly shown by~\autoref{fig:ood-window}: the large-$N$ decay of the ID loss scales with $(NT/(DK))$, while the early-$N$ decay develops a $D$-independent behaviour.
Correspondingly, the OOD losses at early stopping (dashed lines) show an initial descent, followed by a steep ascent around $N\sim (DK)/T$.
This result shows that the transience of ICL is inevitable, unless the learning model is regularised.

\begin{figure}[h!]
  \centering
  \includegraphics[width=\linewidth]{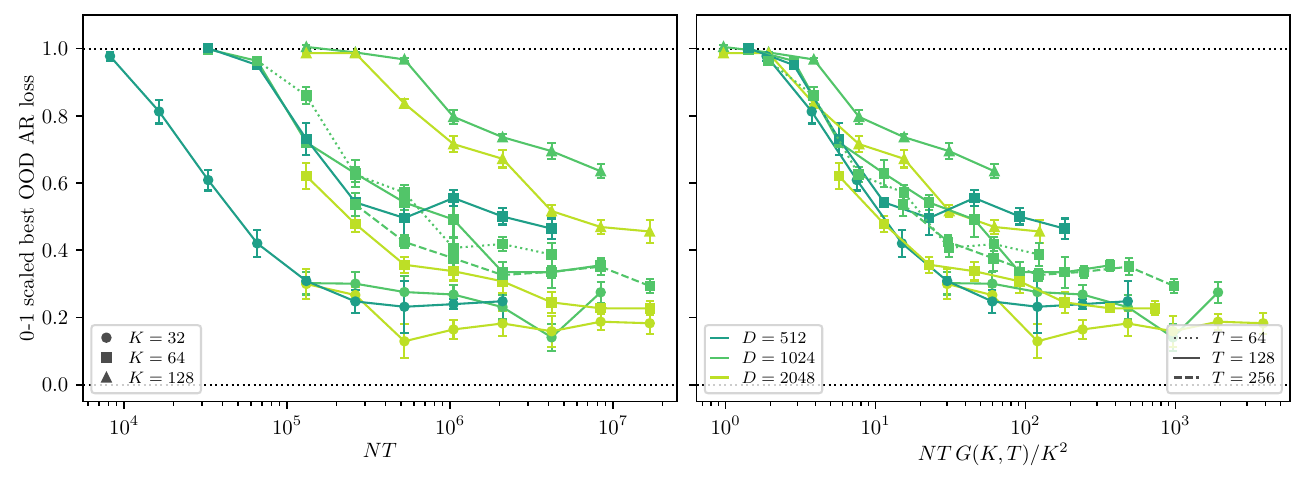}
  \vspace{-10pt}
  \caption{
   \textbf{Empirical OOD learning curves decay at the predicted sample scale.}
   \textbf{Left:} Scaled OOD autoregressive loss $(\mathcal{L}^\star_{\rm AR}(N)-\mathcal{L}^\star_{\rm AR}(\infty))/(\log{K}-\mathcal{L}^\star_{\rm AR}(\infty))$ versus number of training tokens $NT$ for one-block transformers trained on random topic model data with varying $(K,D)$ at $T\,{=}\,128$ (solid), and with $K\,{=}\,64$, $D\,{=}\,1024$ at $T\,{=}\,64$ (dotted) and $T\,{=}\,256$ (dashed).
   These losses are not recorded at early stopping (training step with the lowest ID validation loss), but at the training step with the lowest OOD validation loss.
   Hence, the OOD losses shown in this figure do not rise at $N\sim DK/T$ as in~\autoref{fig:ood-window}.
   \textbf{Right:} The early-$N$ departure from the trivial loss $\log{K}$ does not depend on $D$ and collapse after dividing the number of training sequences by $K^2/G(K,T)$, in perfect agreement with the theory of~\autoref{eq:n-icl}.
  }
  \label{fig:ood-scaling}
\end{figure}

\autoref{fig:ood-scaling} shows OOD learning curves in the high-diversity regime: the initial $D$-independent descent scales like $K^2/G(K,T)$ as predicted in~\autoref{eq:n-icl}.
Since specialisation occurs at $NT\sim DK$, the in-context solution remains stable for a window of training set sizes linear in $D$.
\autoref{fig:id-kscaling} of~\autoref{app:extra-figures} shows the same scaling behaviour for the initial decay of ID learning curves.

\subsection{Comparison with spectral baselines}\label{ssec:spectral}

Latent-variable models like the random topic model are classically learnt by spectral methods based on the moments of the tokens~\citep{hsu2012spectral,balle2014methods,anandkumar2014tensor}.
The first step is recovery of the $D$-dimensional subspace spanned by the topics from a low-order moment: the pair moment $M_2=\frac1D\sum_z\mathbf{p}_z\mathbf{p}_z^\top$, of rank $D$, when $D\leq K$, and the moment of $2m$ tokens when $D\leq\binom{K+m-1}{m}$, the number of independent products of $m$ topic entries.
The topics themselves then follow from the moment with one more token, which is the identifiability step of \autoref{sec:special-theory} (\autoref{app:spectral}).
The subspace is resolved once the empirical moment, estimated from $N$ independent $2m$-tuples, is close to the population moment.
The corresponding sample size is a lower bound for the method to work.

The error on the empirical moment is of order $(K^m N)^{-1/2}$ by the matrix Bernstein inequality (\autoref{app:spectral}, verified numerically in~\autoref{fig:M2-convergence}).
The scales of the population moments, instead, follow from random matrix theory, in the limit $D, K\to\infty$.
With $D/K\,{=}\,\alpha<1$ and $m\,{=}\,1$, the bulk of $M_2$'s spectrum converges to a Marchenko--Pastur distribution with eigenvalues of order $1/(DK)$ (\autoref{fig:M2-spectrum}).
Comparison with the noise gives $N \gtrsim D^2 K$.
With $D/(K(K+1)/2)\,{=}\,\beta<1$ and $m\,{=}\,2$, the relevant bulk of $M_4$ is of order $1/(DK^2)$ (\autoref{fig:M4-spectrum}).
Comparison with the noise gives $N \gtrsim D^2 K^2$.
This analysis is carried out in a different limit from the rest of the paper, and it counts independent tuples rather than documents of $T$ tokens, so we only compare the dependence on task diversity:
quadratic in $D$ for the spectral baseline, against the linear $NT\sim DK$ of the Bayes-optimal learner, which the trained transformers follow.
Furthermore, in the spectral baseline, the order of $K$ grows with the order of the required moment.

\section{Conclusions}

We have studied a solvable model of a mixture of $D$ topics, where two strategies compete: in-distribution specialisation to the finite mixture, or out-of-distribution generalisation to topics never seen in training.
Specialisation requires more training tokens than there are parameters in the data-generating process---a statistical limit that no learner can beat.
The in-context solution is available at a $D$-independent scale, hence it comes first whenever the number of topics is large enough.
Trained transformers follow both scales.
It follows that the transience of ICL~\citep{singh2023transient} is inevitable as the corpus grows, unless a constraint on parameters or training budget keeps the learner in the in-context window.

The specialised solution was called \emph{memorising} in previous studies of the same competition~\citep{raventos2023pretraining,nguyen2025differential,carroll2025dynamics,wurgaft2025incontext}.
Indeed, its placement after the in-context phase mirrors the training dynamics of diffusion models, where generalisation is acquired before the memorisation of the training set~\citep{favero2025bigger}.
We find the label misleading: the specialised predictor stores the $D$ topics of the population, not the $N$ documents of the corpus---memorisation happens in \emph{feature} space.
This is a form of in-distribution generalisation, distinct from the adaptation to new topics of the in-context predictor, and we name it after the specialisation of teacher-student networks, where the hidden units of the student align with those of the teacher~\citep{goldt2019dynamics, barbier2025optimal}.

Several directions are open.
The first is the training dynamics of transformers on the random topic model, which would connect our sample complexities with the phenomenological description of~\citet{wurgaft2025incontext}.
More importantly, the dynamics determine which representations are learned in each phase and the corresponding geometry of the residual stream~\citep{shai2024transformers,ray2026nonergodic}.
Clarifying the distinction between representations in the two phases could help understand the emergence and role of task vectors~\citep{hendel2023incontext}, including scenarios where multiple solutions coexist~\citep{yan2026task}.
The second is the competition between task diversity and other structures of the data.
The tools developed here can address this problem in general non-ergodic hidden Markov models~\citep{riechers2025next}, of which the random topic model is the simplest instance.
Such mixtures are natural models of large heterogeneous corpora.
Solvable models like the one proposed here could help understand how heterogeneity contributes to data statistics and scaling laws~\citep{cagnetta2026deriving}, and how improvements on the individual components of a mixture interact.

\newpage
\subsection*{AI use statement}

In this work, we used generative AI tools for:
assistance in the writing of proofs; implementation of numerical methods.
We have not used generative AI tools for:
Generating synthetic data sets, helping develop theoretical models or conceptual frameworks, formulating mathematical claims, providing critical ingredients for proving mathematical claims, proposing or refining hypotheses, designing feedback on research methodology or experiments, interpreting results,
and assisting with translation, cleaning and reformatting dataset, supporting qualitative and thematic data analysis are not applicable to this work.
Additionally, we used generative AI tools for:
modifying scientific figures, drafting parts of a research paper, identifying relevant literature.
We have reviewed all AI-assisted work.
We checked LLM-generated proofs; LLM-generated numerical implementations were verified and tested for correctness by 2 authors; We checked LLM-generated descriptions of additional related works.
We take responsibility for the final content of this work, including text, claims or artifacts produced with the aid of generative AI.

\subsection*{Reproducibility statement}

The theoretical results are stated with their assumptions and proved in the appendices.
Appendices~\ref{app:topic-stats}--\ref{app:in-context} prove the statements of \autoref{sec:rtm}: the averages over the topic prior, the optimality and loss decomposition of the two predictors, and the exact finite-$t$ loss of the in-context predictor, \autoref{eq:in-context-exact}, from which the in-context reference, the gain $G(K,T)$ and the sample scale $N_{\mathrm{icl}}$ used to rescale the axes of the figures are computed.
\autoref{app:bayes-learner} states the identifiability and regularity assumptions behind the $DK/NT$ law, \autoref{eq:dk-nt-law}, gives sufficient conditions on $(D,K,T)$ under which they hold almost surely, proves the law and gives its fixed-position version.
\autoref{app:icl} lists the conditions under which the estimation excess of a fitted family equals its parameter count, behind \autoref{eq:n-icl}, and \autoref{app:spectral} states the joint limit in which the spectral scales hold.
Arguments that we do not prove are identified as such where they appear: the concentration of pairwise divergences behind the $\log D$ constant, the signal-to-noise criterion and the block approximation of the spectral baseline, and the applicability of the estimation argument to a trained transformer.

The experiments are fully specified in \autoref{app:experiments}: the data-generating process, with the topic prior, the sizes of the validation and OOD sets and the number of fresh OOD topics; the architecture, initialisation and parameter count; the optimiser, learning-rate schedule and batching; the evaluation protocol, the definition of early stopping and the two readings of the OOD loss; and the hyperparameter grid with the selection rule.
Every point of a learning curve is the mean over four runs with independent topic matrix, corpus and initialisation, with standard errors as error bars, and every run stores its full configuration, including the seeds of the draws, so that any point can be regenerated exactly.
The theoretical references in the figures are evaluated from closed-form expressions or by Monte Carlo estimates over topic realisations, as stated in the captions.
The code for data generation, training, evaluation and the theoretical curves is available at \url{https://github.com/fracagnetta/random-topic-model}.

\newpage
\bibliography{iclbeforespecial}
\bibliographystyle{iclr2027_conference}
\newpage

\appendix
\section{Statistics over random topic models}\label{app:topic-stats}

This appendix collects the averages over the topic prior used in the main text.
Let $\mathbf{p},\mathbf{p}'\sim\mathrm{Dir}_K(\mathbf{1})$ be independent, i.e. uniform on the simplex $\Delta^{K-1}$, and let $\psi\,{=}\,\Gamma'/\Gamma$ denote the digamma function.
Each entry is marginally $p_a\sim\mathrm{Beta}(1,K{-}1)$, with moments
\begin{align}\label{eq:dirichlet-moments}
  \overline{p_a^{\,r}} = \frac{\Gamma(K)\,\Gamma(r{+}1)}{\Gamma(K{+}r)},
  \qquad
  \overline{\log p_a} = \psi(1) - \psi(K),
  \qquad
  \overline{p_a\log p_a} = \frac{1}{K}\left[\psi(2) - \psi(K{+}1)\right],
\end{align}
where the last two follow by differentiating the first with respect to $r$ at $r\,{=}\,0$ and $r\,{=}\,1$.

\paragraph{Entropy of a topic.}
By symmetry of the entries,
\begin{align}\label{eq:dirichlet-entropy}
  \overline{H(\mathbf{p})} = -K\,\overline{p_a\log p_a} = \psi(K{+}1) - \psi(2)
  = \log K - 1 + \gamma_{\mathrm{E}} + \frac{1}{2K} + O\!\left(K^{-2}\right),
\end{align}
using $\psi(2)\,{=}\,1-\gamma_{\mathrm{E}}$ and $\psi(K{+}1)\,{=}\,\log K + 1/(2K) + O(K^{-2})$.
A random topic thus has entropy $1-\gamma_{\mathrm{E}}-1/(2K)\approx0.42$ nats below the uniform value $\log K$, at leading orders: this gap is the height of the descent of every profile in~\autoref{fig:id-vs-ood-profiles}.

\paragraph{Divergence between two topics.}
Expanding the divergence and using the independence of $\mathbf{p}$ and $\mathbf{p}'$,
\begin{align}\label{eq:dirichlet-kl}
  \overline{\KL(\mathbf{p}\Vert\mathbf{p}')}
  = K\,\overline{p_a\log p_a} - K\,\overline{p_a}\;\overline{\log p'_a}
  = \left[\psi(2)-\psi(K{+}1)\right] - \left[\psi(1)-\psi(K)\right] = 1 - \frac{1}{K},
\end{align}
by the recurrence $\psi(x{+}1)\,{=}\,\psi(x)+1/x$.
Two random topics are thus separated by an order-one divergence for every vocabulary size, which sets the rate of topic resolution in~\autoref{app:specialised}.

\paragraph{Bhattacharyya coefficient.}
The non-asymptotic resolution bound of~\autoref{app:specialised} involves $\rho(\mathbf{p},\mathbf{p}')\,{:=}\,\sum_a\sqrt{p_a p'_a}$, whose average follows from~\autoref{eq:dirichlet-moments} at $r\,{=}\,1/2$,
\begin{align}\label{eq:dirichlet-bhattacharyya}
  \overline{\rho(\mathbf{p},\mathbf{p}')} = K\left(\overline{\sqrt{p_a}}\right)^2
  = \frac{\pi K}{4}\left[\frac{\Gamma(K)}{\Gamma(K{+}\tfrac12)}\right]^2
  = \frac{\pi}{4}\left(1 + \frac{1}{4K} + O\!\left(K^{-2}\right)\right).
\end{align}

\section{The specialised predictor}\label{app:specialised}

This appendix proves the statements of~\autoref{ssec:specialised}: the optimality of $\mathbf{q}_t$, the loss decomposition~\autoref{eq:specialised-loss}, and the resolution of the topic on the context scale $\log D$.
Throughout, the topic matrix $P$ is fixed, with distinct columns, and all expectations are over documents at fixed $P$.

\paragraph{Optimality.}
Conditional on the context, $X_{t+1}$ is distributed as $\mathbf{q}_t(X_{\leq t})$: by construction $q_t(a\mid X_{\leq t}) = \sum_z \mathbb{P}(Z\,{=}\,z\mid X_{\leq t})\,\mathbb{P}(X_{t+1}\,{=}\,a\mid Z\,{=}\,z)$, using that $X_{t+1}$ is independent of the context given $Z$.
For any predictor $\mathbf{f}$, therefore,
\begin{align}\label{eq:specialised-optimality}
  \mathbb{E}\left[-\log f_t(X_{t+1}\mid X_{\leq t}) \,\middle|\, X_{\leq t}\right]
  = H(\mathbf{q}_t) + \KL(\mathbf{q}_t\Vert\mathbf{f}_t) \geq H(\mathbf{q}_t),
\end{align}
with equality if and only if $\mathbf{f}_t\,{=}\,\mathbf{q}_t$.
Averaging over contexts, $\mathcal{L}_t[\mathbf{f}]\geq\mathcal{L}_t[\mathbf{q}]$.

\paragraph{Loss decomposition.}
Conditioning instead on the topic and the context, $X_{t+1}\sim\mathbf{p}_Z$, and the same identity gives $\mathbb{E}[-\log q_t(X_{t+1}\mid X_{\leq t})\mid Z, X_{\leq t}] = H(\mathbf{p}_Z) + \KL(\mathbf{p}_Z\Vert\mathbf{q}_t)$; averaging over the topic, uniform on $\lbrace 1,\dots,D\rbrace$, and over the context yields~\autoref{eq:specialised-loss}.
The floor is the loss of the oracle $\mathbf{f}_t\,{=}\,\mathbf{p}_Z$, which is not a function of the context and hence not a predictor in the sense of~\autoref{eq:ntp-loss}; by the optimality of $\mathbf{q}_t$, every predictor pays at least the excess $\mathbb{E}[\KL(\mathbf{p}_Z\Vert\mathbf{q}_t)]$ above it.

\paragraph{Excess loss and wrong-topic mass.}
Write the posterior in terms of the likelihood ratios of the wrong topics against the true one,
\begin{align}\label{eq:likelihood-ratio}
  R^{(t)}_{y} := \frac{\gamma_t(y)}{\gamma_t(Z)} = \prod_{s=1}^t \frac{p_y(X_s)}{p_Z(X_s)},
  \qquad
  \gamma_t(Z) = \Big(1 + \sum_{y\neq Z} R^{(t)}_y\Big)^{-1}.
\end{align}
Isolating the true topic in~\autoref{eq:specialised}, $\mathbf{q}_t = \gamma_t(Z)\,\mathbf{p}_Z + \sum_{y\neq Z}\gamma_t(y)\,\mathbf{p}_y \geq \gamma_t(Z)\,\mathbf{p}_Z$ entrywise, hence
\begin{align}\label{eq:excess-bound}
  \KL(\mathbf{p}_Z\Vert\mathbf{q}_t) \leq \sum_{a=1}^K p_Z(a)\log\frac{p_Z(a)}{\gamma_t(Z)\,p_Z(a)} = -\log\gamma_t(Z) = \log\Big(1 + \sum_{y\neq Z} R^{(t)}_y\Big).
\end{align}
Conditional on $Z$, the tokens are i.i.d. from $\mathbf{p}_Z$, so by the strong law of large numbers
\begin{align}\label{eq:lln-rate}
  \frac{1}{t}\log R^{(t)}_y = \frac{1}{t}\sum_{s=1}^t \log\frac{p_y(X_s)}{p_Z(X_s)} \xrightarrow[t\to\infty]{\mathrm{a.s.}} -\KL(\mathbf{p}_Z\Vert\mathbf{p}_y) < 0,
\end{align}
i.e. each wrong topic is suppressed exponentially in $t$, at a rate equal to its divergence from the true topic.
This is the mechanism described in the main text; it is asymptotic in $t$ at fixed $P$.

\paragraph{Resolution on the scale $\log D$.}
A bound valid at every $t$ follows from a fractional moment of the likelihood ratios.
Using $\log(1+\sum_y R_y)\leq\sum_y\log(1+R_y)\leq\sum_y\sqrt{R_y}$ in~\autoref{eq:excess-bound}, then the independence of the tokens given $Z$,
\begin{align}\label{eq:bhattacharyya-bound}
  \mathbb{E}\left[\KL(\mathbf{p}_Z\Vert\mathbf{q}_t)\right] \leq \mathbb{E}\left[-\log\gamma_t(Z)\right] &\leq \frac{1}{D}\sum_{z=1}^D\sum_{y\neq z} \mathbb{E}\Big[\sqrt{R^{(t)}_y}\,\Big|\,Z\,{=}\,z\Big] \nonumber\\
  & = \frac{1}{D}\sum_{z=1}^D\sum_{y\neq z} \rho_{zy}^{\,t} \leq (D\,{-}\,1)\,\rho_\star^{\,t},
\end{align}
where $\rho_{zy}\,{:=}\,\sum_a\sqrt{p_z(a)\,p_y(a)}$ is the Bhattacharyya coefficient of topics $z$ and $y$, and $\rho_\star\,{:=}\,\max_{z\neq y}\rho_{zy}\,{<}\,1$ by the Cauchy--Schwarz inequality and the distinctness of the topics.
Hence the excess loss of the specialised predictor is below any $\epsilon\,{>}\,0$ as soon as
\begin{align}\label{eq:resolution-scale}
  t \geq \frac{\log\left[(D-1)/\epsilon\right]}{\log(1/\rho_\star)},
\end{align}
which is the $\log D$ scale of~\autoref{ssec:specialised}, with a rate set by the closest pair of topics.
The guaranteed rate $\log(1/\rho_{zy})$ is at most half of the almost-sure rate in~\autoref{eq:lln-rate}, since $\log\rho_{zy}\geq-\tfrac12\KL(\mathbf{p}_z\Vert\mathbf{p}_y)$ by Jensen's inequality, and $\rho_\star$ grows with $D$ as the closest pair among $D$ random topics gets closer: \autoref{eq:resolution-scale} is a statement about the dependence on $t$ at fixed $P$, not a sharp constant.
For the constant, the typical pair is the relevant one: two random topics have $\overline{\KL(\mathbf{p}\Vert\mathbf{p}')}\,{=}\,1-1/K$ and $\overline{\rho(\mathbf{p},\mathbf{p}')}\,{\simeq}\,\pi/4$ (\autoref{app:topic-stats}), and if the $D\,{-}\,1$ divergences from the true topic concentrate near their mean, the odds of a wrong topic in~\autoref{eq:excess-bound} are approximately $(D\,{-}\,1)\,e^{-(1-1/K)\,t}$, so that the floor is reached for $t\approx\log(D\,{-}\,1)$.
This last estimate is heuristic, as it assumes a concentration of the pairwise divergences that we do not prove; it is consistent with the half-descent points in~\autoref{fig:id-vs-ood-profiles}, which grow linearly in $\log D$.

\section{The in-context predictor}\label{app:in-context}

This appendix proves the statements of~\autoref{ssec:in-context}: the form and optimality of $\mathbf{r}_t$, the loss decomposition~\autoref{eq:in-context-loss}, and the exact finite-$t$ risk behind its expansion.
Throughout, the document is OOD: $\mathbf{p}_\star\sim\mathrm{Dir}_K(\mathbf{1})$ and, given $\mathbf{p}_\star$, the tokens are i.i.d. from $\mathbf{p}_\star$.

\paragraph{Posterior and add-one rule.}
The prior density is constant on the simplex and the likelihood of a context is $\prod_a p_\star(a)^{C_a}$, so the posterior density is proportional to $\prod_a p_\star(a)^{C_a}$: this is $\mathrm{Dir}_K(C_1{+}1,\dots,C_K{+}1)$, whose density is proportional to $\prod_a p_a^{\alpha_a-1}$ with $\alpha_a\,{=}\,C_a{+}1$.
Given $\mathbf{p}_\star$ and the context, $X_{t+1}\sim\mathbf{p}_\star$, and the mean of $\mathrm{Dir}_K(\boldsymbol{\alpha})$ is $\alpha_a/\sum_b\alpha_b$, so the posterior predictive distribution is
\begin{align}\label{eq:add-one-derivation}
  r_t(a\mid X_{\leq t}) = \mathbb{E}\left[p_\star(a)\mid X_{\leq t}\right] = \frac{C_a+1}{t+K},
\end{align}
which is~\autoref{eq:in-context}.

\paragraph{Optimality.}
As in~\autoref{app:specialised}, $\mathbf{r}_t$ is the conditional distribution of $X_{t+1}$ given the context, so for any predictor $\mathbf{f}$
\begin{align}\label{eq:in-context-optimality}
  \mathbb{E}\left[-\log f_t(X_{t+1}\mid X_{\leq t}) \,\middle|\, X_{\leq t}\right]
  = H(\mathbf{r}_t) + \KL(\mathbf{r}_t\Vert\mathbf{f}_t) \geq H(\mathbf{r}_t),
\end{align}
with equality if and only if $\mathbf{f}_t\,{=}\,\mathbf{r}_t$; averaging over contexts, $\mathcal{L}^\star_t[\mathbf{f}]\geq\mathcal{L}^\star_t[\mathbf{r}]$.
Unlike $\mathbf{q}_t$, the predictor $\mathbf{r}_t$ does not depend on the topic matrix: on a document generated from any topic $\mathbf{p}$, its loss is a function $R_t(\mathbf{p})\,{:=}\,\mathbb{E}[-\log r_t(X_{t+1}\mid X_{\leq t})\mid\mathbf{p}]$ of that topic alone.
Its ID loss at fixed $P$ is $\mathcal{L}_t[\mathbf{r}]\,{=}\,\frac1D\sum_z R_t(\mathbf{p}_z)$ and its OOD loss is $\mathcal{L}^\star_t[\mathbf{r}]\,{=}\,\overline{R_t(\mathbf{p})}$, whence $\overline{\mathcal{L}_t[\mathbf{r}]}\,{=}\,\mathcal{L}^\star_t[\mathbf{r}]$.

\paragraph{Loss decomposition.}
Conditioning on $\mathbf{p}_\star$ and the context, $X_{t+1}\sim\mathbf{p}_\star$, so $\mathbb{E}[-\log r_t(X_{t+1}\mid X_{\leq t})\mid\mathbf{p}_\star,X_{\leq t}] = H(\mathbf{p}_\star)+\KL(\mathbf{p}_\star\Vert\mathbf{r}_t)$; averaging over $\mathbf{p}_\star$ and the context gives the first equality in~\autoref{eq:in-context-loss}, with $\mathbb{E}[H(\mathbf{p}_\star)]\,{=}\,\overline{H(\mathbf{p})}$ because $\mathbf{p}_\star$ has the law of the training topics.
Moreover, by~\autoref{eq:in-context-optimality} at $\mathbf{f}\,{=}\,\mathbf{r}$, $\mathcal{L}^\star_t[\mathbf{r}]\,{=}\,H(X_{t+1}\mid X_{\leq t})$ is a conditional entropy; since the tokens are exchangeable given $\mathbf{p}_\star$, $H(X_{t+2}\mid X_{\leq t})\,{=}\,H(X_{t+1}\mid X_{\leq t})$, and conditioning on one more token shows that $\mathcal{L}^\star_t[\mathbf{r}]$ is non-increasing in $t$.

\paragraph{Exact finite-$t$ risk.}
Integrating the likelihood against the uniform prior, the count vector of a context is uniform over the $\binom{t+K-1}{K-1}$ weak compositions of $t$ into $K$ parts,
\begin{align}\label{eq:counts-uniform}
  \mathbb{P}\left(C(X_{\leq t})\,{=}\,\mathbf{n}\right)
  = \frac{t!}{\prod_a n_a!}\;\frac{\Gamma(K)\prod_a\Gamma(n_a{+}1)}{\Gamma(t{+}K)}
  = \binom{t+K-1}{K-1}^{-1},
  \qquad \textstyle\sum_a n_a = t.
\end{align}
Hence, by~\autoref{eq:add-one-derivation} and the symmetry of the entries, for $K\geq2$
\begin{align}\label{eq:in-context-exact}
  \mathcal{L}^\star_t[\mathbf{r}]
  &= \log(t+K) - \mathbb{E}\left[\sum_a r_t(a\mid X_{\leq t})\log(C_a+1)\right] \nonumber\\
  &= \log(t+K) - \frac{K}{(t+K)\binom{t+K-1}{K-1}}\sum_{n=0}^{t}(n+1)\log(n+1)\binom{t-n+K-2}{K-2},
\end{align}
where $\binom{t-n+K-2}{K-2}$ counts the compositions with $C_1\,{=}\,n$.
In particular $\mathcal{L}^\star_0[\mathbf{r}]\,{=}\,\log K$ and $\mathcal{L}^\star_t[\mathbf{r}]\to\overline{H(\mathbf{p})}$ as $t\to\infty$.
Expanding at large $t$ and fixed $K$ gives the second equality in~\autoref{eq:in-context-loss},
\begin{align}\label{eq:in-context-expansion}
  \mathcal{L}^\star_t[\mathbf{r}] = \psi(K{+}1)-\psi(2) + \frac{K-1}{2t} + O_K\!\left(\frac{\log t}{t^2}\right),
\end{align}
where the logarithm in the remainder comes from the mass that the uniform prior puts near the boundary of the simplex.
The leading correction is one half of the number of parameters of a topic, $K\,{-}\,1$, per observation, and the expansion requires $t\gg K$: at $K\,{=}\,64$ and $t\,{=}\,128$, \autoref{eq:in-context-exact} gives an excess of $0.15$ nats over the floor, against $0.25$ for the leading term alone; the former is the gap between the dashed line and the floor at the end of the profiles in~\autoref{fig:id-vs-ood-profiles}.

\section{The Bayesian learner}\label{app:bayes-learner}

This appendix proves the statements of \autoref{sec:special-theory}: the form and optimality of the Bayesian learner, the information identity, the $DK/NT$ law and its fixed-position version.

\paragraph{Posterior and predictor.}
Let $\mathcal{C}_N = (X^{(n)}_{1:T})_{n=1}^N$ be the corpus, with count vectors $C^{(n)}$, and write $\ell_C(\mathbf{p}) := \prod_a p(a)^{C_a}$ for the likelihood of a count vector under a topic.
Given a candidate topic matrix $\tilde P$, integrating out the latent topic of each document gives the likelihood of corpus and context,
\begin{align}\label{eq:corpus-likelihood}
  L(\tilde P) := \prod_{n=1}^N \Big[\frac{1}{D}\sum_{z=1}^D \ell_{C^{(n)}}(\tilde{\mathbf{p}}_z)\Big] \times \frac{1}{D}\sum_{z=1}^D \ell_{C(X_{\leq t})}(\tilde{\mathbf{p}}_z),
\end{align}
up to multinomial factors that do not depend on $\tilde P$.
The posterior of $P$ is $\pi(\mathrm{d}\tilde P) \propto L(\tilde P)\,\mu(\mathrm{d}\tilde P)$, with $\mu$ the product of $D$ uniform priors on the simplex, and \autoref{eq:bayes-learner} reads $\hat{\mathbf{q}}_t = \int \mathbf{q}_{t,\tilde P}(X_{\leq t})\,\pi(\mathrm{d}\tilde P)$, where $\mathbf{q}_{t,\tilde P}$ is the specialised predictor of \autoref{eq:specialised} computed with $\tilde P$.
By construction $\hat{\mathbf{q}}_t$ is the conditional law of $X_{t+1}$ given $(\mathcal{C}_N, X_{\leq t})$ under the joint law of topics, corpus and test document, so, as in \autoref{app:specialised}, every predictor $\mathbf{f}$ built from corpus and context satisfies $\overline{\mathcal{L}_t[\mathbf{f}]} \geq \overline{\mathcal{L}_t[\hat{\mathbf{q}}]}$, with equality if and only if $\mathbf{f}_t\,{=}\,\hat{\mathbf{q}}_t$.

\paragraph{Excess and mutual information.}
At fixed $P$, conditioning on the context, $X_{t+1}\sim\mathbf{q}_t$ and the corpus is independent of the test document, so $\mathcal{L}_t[\hat{\mathbf{q}}] - \mathcal{L}_t[\mathbf{q}] = \mathbb{E}[\KL(\mathbf{q}_t\Vert\hat{\mathbf{q}}_t)] \geq 0$.
Averaged over $P$, the two losses are the conditional entropies $H(X_{t+1}\mid X_{\leq t},\mathcal{C}_N)$ and $H(X_{t+1}\mid X_{\leq t},\mathcal{C}_N,P)$, whence
\begin{align}\label{eq:excess-mi}
  \varepsilon_t(N) = I(P; X_{t+1}\mid X_{\leq t}, \mathcal{C}_N).
\end{align}
The second entropy does not change if one more document $X^{(N+1)}_{1:T}$ is added to the corpus, since given $P$ the corpus is irrelevant, so $\varepsilon_t(N)-\varepsilon_t(N+1) = I(X_{t+1}; X^{(N+1)}_{1:T}\mid X_{\leq t},\mathcal{C}_N)\geq0$: the learning curve is non-increasing in $N$.
Summing \autoref{eq:excess-mi} over positions with the chain rule, and using that the test document has the law of $X^{(N+1)}_{1:T}$ and is independent of the corpus given $P$,
\begin{align}\label{eq:excess-chain}
  \sum_{t=0}^{T-1}\varepsilon_t(N) = I\big(P; X^{(N+1)}_{1:T}\,\big|\,\mathcal{C}_N\big) = I(P;\mathcal{C}_{N+1}) - I(P;\mathcal{C}_N),
\end{align}
which is \autoref{eq:learning-curve-information}, the document-level form of the cumulative predictive risk identity of \citet{clarke1990information}.

\paragraph{The $DK/NT$ law.}
Up to a relabelling of the topics, the law $Q_P := \frac{1}{D}\sum_z \mathbf{p}_z^{\otimes T}$ of a document has $d = D(K-1)$ parameters.
Assume that $P\mapsto Q_P$ is identifiable up to relabelling and regular: twice continuously differentiable in local coordinates, with positive definite Fisher information, and with a prior density that is positive and continuous.
Then, for fixed $D$, $K$, $T$ and $N\to\infty$, Theorem 1 of \citet{clarke1994jeffreys} gives
\begin{align}\label{eq:clarke-barron}
  I(P;\mathcal{C}_N) = \frac{d}{2}\log N + c + o(1),
\end{align}
with a constant $c$ that depends on prior and document model but not on $N$.
The increment in \autoref{eq:excess-chain} does not follow by direct subtraction, since an $o(1)$ term need not have $o(N^{-1})$ increments, but it does follow from monotonicity.
Let $a_N := \sum_t \varepsilon_t(N)$, non-increasing in $N$, and for $\lambda>1$ let $m=\lfloor\lambda N\rfloor$ and $l=\lfloor N/\lambda\rfloor$.
Summing $a_n$ over $N\leq n<m$ and over $l\leq n<N$ and using monotonicity,
\begin{align}\label{eq:sandwich}
  \frac{I(P;\mathcal{C}_m)-I(P;\mathcal{C}_N)}{m-N} \;\leq\; a_N \;\leq\; \frac{I(P;\mathcal{C}_N)-I(P;\mathcal{C}_l)}{N-l},
\end{align}
where both numerators equal $\frac{d}{2}\log\lambda+o(1)$ by \autoref{eq:clarke-barron}.
Letting $N\to\infty$ and then $\lambda\downarrow1$ squeezes $N a_N$ between $\frac{d}{2}\frac{\log\lambda}{\lambda-1}$ and $\frac{d}{2}\frac{\lambda\log\lambda}{\lambda-1}$, both of which tend to $d/2$; dividing by $T$ gives \autoref{eq:dk-nt-law}.

\paragraph{Identifiability.}
For $T\geq3$ the third-order moment of the tokens, $M_3 = \frac{1}{D}\sum_z \mathbf{p}_z^{\otimes3}$, is a marginal of $Q_P$, and by Kruskal's theorem~\citep{kruskal1977threeway,allman2009identifiability} its decomposition into $D$ rank-one terms is unique up to permutation when $3\,\mathrm{rank}_{\mathrm{K}}(P) \geq 2D+2$, where the Kruskal rank $\mathrm{rank}_{\mathrm{K}}(P)$ is the largest $k$ such that any $k$ columns of $P$ are linearly independent.
For topics drawn from the simplex, $\mathrm{rank}_{\mathrm{K}}(P)=\min\{K,D\}$ almost surely, so the condition holds for $D\leq K$ (and up to $D\leq 3K/2-1$), and the topics can be recovered constructively: the second moment $M_2=\frac1D\sum_z\mathbf{p}_z\mathbf{p}_z^\top$ has rank $D$ and whitens $M_3$ into an orthogonally decomposable tensor whose factors are the topics~\citep{anandkumar2014tensor}.
For larger $D$, lift the topics to $\mathbf{v}_z=\mathbf{p}_z^{\otimes m}$, which lie in the space of symmetric tensors, of dimension $s_m=\binom{K+m-1}{m}$, and span it as $\mathbf{p}_z$ ranges over the simplex, since a homogeneous polynomial vanishing on the simplex vanishes identically; some $s_m\times s_m$ minor of $[\mathbf{v}_1,\dots,\mathbf{v}_D]$ is then a nonzero polynomial in the topic entries, so the lifted topics are linearly independent almost surely whenever $D\leq s_m$.
In that case the $2m$-token moment $\frac1D\sum_z\mathbf{v}_z\mathbf{v}_z^\top$ has rank $D$, the rank of the Hankel matrix of the token process, and together with the $(2m{+}1)$-token moment it determines the law of documents of every length, hence $P$ up to relabelling~\citep{balle2014methods} (\autoref{app:spectral}).
Moreover, for $T\geq2m+1$ the Fisher information of a document is nonsingular: a perturbation $\delta P$ that leaves the law of a document unchanged to first order leaves the $(2m{+}1)$-token marginal $\frac1D\sum_z\mathbf{v}_z\otimes\mathbf{v}_z\otimes\mathbf{p}_z$ unchanged, and contracting its variation with $a_z\otimes a_z\otimes\mathrm{Id}$, where the functionals $a_z$ satisfy $a_z(\mathbf{v}_{z'})=\delta_{zz'}$, gives $\delta\mathbf{p}_z+2a_z(\delta\mathbf{v}_z)\,\mathbf{p}_z=0$, whose sum over coordinates forces $a_z(\delta\mathbf{v}_z)=0$ and then $\delta\mathbf{p}_z=0$.
Hence $D\leq\binom{K+m-1}{m}$ and $T\geq2m+1$ suffice almost surely: $D\leq K$ with $T\geq3$, $D\leq K(K{+}1)/2$ with $T\geq5$, and so on; these conditions are sufficient, not necessary, and they are the only place where the genericity of the topic matrix is used.
Regularity additionally excludes the sets, of prior measure zero, where two topics coincide or a topic lies on the boundary of the simplex.

\paragraph{Fixed position: the resolved learner.}
Reveal to the learner the topic labels of the training documents and of the test document.
Conditional on the labels, by the conjugacy of \autoref{app:in-context}, the Bayesian predictor is the add-one rule on the pooled counts of the test topic,
\begin{align}\label{eq:resolved-predictor}
  \hat{q}^{\mathrm{res}}_t(a\mid X_{\leq t}) = \frac{M_{Z,a} + C_a + 1}{N_Z T + t + K},
\end{align}
where $N_Z$ is the number of training documents of topic $Z$ and $M_{Z,a}$ their total count of token $a$: the in-context predictor of \autoref{eq:in-context} applied to $m = N_ZT+t$ tokens.
Averaged over topics, its excess over the entropy rate $\frac1D\sum_z H(\mathbf{p}_z)$ is therefore $\mathbb{E}[g_K(N_ZT+t)]$, with $g_K(m) := \mathcal{L}^\star_m[\mathbf{r}] - \overline{H(\mathbf{p})}$ the excess of the in-context predictor in \autoref{eq:in-context-loss} and $N_Z\sim\mathrm{Bin}(N,1/D)$.
If $N/D\to\infty$ then $N_Z/(N/D)\to1$ in probability, and if moreover $tD/(NT)\to0$ and $\log T=o(N/D)$, so that the event that the test topic is absent from the corpus is negligible on the $1/(NT)$ scale, \autoref{eq:in-context-loss} gives
\begin{align}\label{eq:resolved-learning-curve}
  \varepsilon^{\mathrm{res}}_t(N) := \mathbb{E}\left[g_K(N_ZT+t)\right] = \frac{D(K-1)}{2NT}\,\big(1+o(1)\big).
\end{align}
The resolved learner is not the Bayesian learner, which must also infer the labels and the topic matrix.
For a fixed $P$ with distinct topics, and assuming identifiability from a marginal of $k\leq T$ tokens, the two excesses differ by at most
\begin{align}\label{eq:transfer-bound}
  C(P)\left(e^{-c(P)\,t} + N e^{-c(P)\,T}\right) + \exp\big\{-a(P)\,N + O_P\big(\log[N(T+t)]\big)\big\},
\end{align}
with $c(P) = \log(1/\rho_\star)$ from \autoref{eq:bhattacharyya-bound} and $a(P)>0$.
The first two terms are the costs of unresolved test and training labels, by the bound of \autoref{app:specialised} applied to the test context and to each of the $N$ training documents; the last is the posterior mass on topic matrices outside a fixed neighbourhood of $P$, exponentially small in $N$ by a likelihood-ratio test on the identifying marginal, against the polynomial prior mass of the neighbourhood.
Hence the Bayesian learner obeys \autoref{eq:resolved-learning-curve} at fixed $P$ whenever $t$ and $T$ grow like $A\log N$ with $A$ large enough for \autoref{eq:transfer-bound} to be $o(1/(NT))$: the fixed-position form of the $DK/NT$ law.
The statement is conditional on $P$; averaging it over the prior would require uniform control over nearly coinciding topics, for which $c(P)$ is arbitrarily small.

\section{The in-context solution}\label{app:icl}

This appendix supports \autoref{sec:icl-theory}: the estimation excess of a fitted parametric family, the count-linear family that contains the in-context predictor, and the gain $G(K,T)$ with its interpretation and expansions.

\paragraph{Estimation excess of a fitted family.}
Let $\{f_\theta:\theta\in\Theta\subset\mathbb{R}^d\}$ be a family of distributions of a document $Y=X_{1:T}$, with all $d$ coordinates of $\theta$ free, and let $\ell(\theta;Y):=-\log f_\theta(Y)$ and $\mathcal{R}(\theta):=\mathbb{E}[\ell(\theta;Y)]$ be its log loss and its population risk under the true law of $Y$.
Let $\theta_\star$ be an interior minimiser of $\mathcal{R}$, isolated up to finite symmetries, around which $f_\theta(y)>0$ is three times continuously differentiable, and define the risk curvature and the gradient covariance
\begin{align}\label{eq:HJ}
  \mathsf{H} := \nabla^2\mathcal{R}(\theta_\star)\succ0,
  \qquad
  \mathsf{J} := \mathbb{E}\left[\nabla\ell(\theta_\star;Y)\,\nabla\ell(\theta_\star;Y)^\top\right];
\end{align}
the gradient has zero mean because $\theta_\star$ is a minimiser.
Let $\hat\theta_N$ be a consistent fit on $N$ independent documents whose empirical gradient is $o_{\mathbb{P}}(N^{-1/2})$, as for an interior minimiser of the empirical risk $\widehat{\mathcal{R}}_N$, and let $g_N:=\nabla\widehat{\mathcal{R}}_N(\theta_\star)$, which has mean zero and covariance $\mathsf{J}/N$ by independence.
Expanding the empirical gradient and the population risk around $\theta_\star$,
\begin{align}\label{eq:two-expansions}
  \hat\theta_N-\theta_\star = -\mathsf{H}^{-1} g_N + u_N,
  \qquad
  \mathcal{R}(\hat\theta_N)-\mathcal{R}(\theta_\star) = \tfrac12(\hat\theta_N-\theta_\star)^\top\mathsf{H}\,(\hat\theta_N-\theta_\star) + v_N,
\end{align}
with $u_N=o_{\mathbb{P}}(N^{-1/2})$ and $v_N=o_{\mathbb{P}}(N^{-1})$.
If the remainders are also small in expectation, $\mathbb{E}\|u_N\|^2=o(N^{-1})$ and $\mathbb{E}|v_N|=o(N^{-1})$, then substituting and using $\mathbb{E}[g^\top A g]=\operatorname{tr}(A\operatorname{Cov}g)$ for centred $g$,
\begin{align}\label{eq:effective-dimension}
  \mathbb{E}\left[\mathcal{R}(\hat\theta_N)-\mathcal{R}(\theta_\star)\right] = \frac{d_{\mathrm{eff}}}{2N} + o(N^{-1}),
  \qquad d_{\mathrm{eff}} := \operatorname{tr}\left(\mathsf{J}\mathsf{H}^{-1}\right).
\end{align}
If the family is well specified, i.e. the true law is $f_{\theta_\star}$, differentiating $\sum_y f_\theta(y)=1$ twice gives $\mathsf{H}=\mathsf{J}$, the Fisher information of one document, and $d_{\mathrm{eff}}=d$ whenever it is nonsingular.
The remainder bounds hold, for instance, for an empirical-risk minimiser over a compact set that contains $\theta_\star$ in its interior and on which probabilities are bounded below; they are not automatic for an unrestricted fit, nor for a transformer.
Divided by $T$, \autoref{eq:effective-dimension} is a per-token excess $d/(2NT)$.
For the topic matrix itself, $d=D(K-1)$, and the Fisher information is nonsingular almost surely for $D\leq K$ and $T\geq3$ (the Jacobian of the three-token marginal is injective when the topics are linearly independent), so a fitted dictionary recovers \autoref{eq:dk-nt-law} by an argument on a fitted parameter rather than on the posterior; for the count map below, $d\simeq K^2$.

\paragraph{The count-linear family.}
The predictors that are linear in the counts up to normalisation,
\begin{align}\label{eq:count-linear-family}
  f_{A,\mathbf{b}}(a\mid X_{\leq t}) := \frac{\big(A\,C(X_{\leq t})+\mathbf{b}\big)_a}{\mathbf{1}^\top\big(A\,C(X_{\leq t})+\mathbf{b}\big)},
  \qquad A\in\mathbb{R}^{K\times K},\ \mathbf{b}\in\mathbb{R}^K,
\end{align}
contain the in-context predictor at $(A,\mathbf{b})=(I,\mathbf{1})$.
Counts are bounded by $T$, so the family is positive and smooth on a neighbourhood of this point, and it has $d=K^2+K-1$ free parameters, the overall scale of $(A,\mathbf{b})$ being immaterial.
On OOD documents the family is well specified, since the law of an OOD document is the product of its add-one predictors (\autoref{app:in-context}), so $\theta_\star=(I,\mathbf{1})$ and \autoref{eq:effective-dimension} gives the excess $d/(2NT)\simeq K^2/(2NT)$ of \autoref{sec:icl-theory} for a learner that fits this family.
On the training distribution the family is misspecified at finite $D$, its optimum being a fit to $\mathbf{q}_t$, a ratio of exponentials of the counts.
The misspecification vanishes as $D\to\infty$: writing \autoref{eq:specialised} as a ratio of averages over the dictionary,
\begin{align}\label{eq:q-to-r}
  q_t(a\mid X_{\leq t}) = \frac{\frac1D\sum_{z} p_z(a)\prod_b p_z(b)^{C_b}}{\frac1D\sum_{z}\prod_b p_z(b)^{C_b}}
  \xrightarrow[D\to\infty]{\mathrm{a.s.}} \frac{\overline{p(a)\prod_b p(b)^{C_b}}}{\overline{\prod_b p(b)^{C_b}}} = \frac{C_a+1}{t+K} = r_t(a\mid X_{\leq t}),
\end{align}
by the strong law of large numbers applied to numerator and denominator at fixed $K$ and count vector, and the Dirichlet moment $\overline{\prod_b p(b)^{n_b}} = \Gamma(K)\prod_b\Gamma(n_b{+}1)/\Gamma(K{+}\textstyle\sum_b n_b)$, of which \autoref{eq:dirichlet-moments} is the one-entry case.
The convergence is pointwise in the count vector, and the dictionary size needed for a given accuracy grows with $t$, since the weights in \autoref{eq:q-to-r} concentrate on a neighbourhood of the empirical frequencies that shrinks with $t$.
Note that the family cannot be reached through the specialised predictor of a small dictionary.
 $\mathbf{q}_t$ is a convex combination of the $D$ topics, so its Jacobian with respect to the counts has rank at most $D-1$, whereas that of the add-one map is $K-1$: a learner whose predictor tracks the in-distribution optimum can realise the count-linear rule only if $D\geq K$.

\paragraph{The gain of the in-context predictor.}
On an OOD document every token is marginally uniform, $\mathbb{P}(X_{t+1}\,{=}\,a)=\overline{p(a)}=1/K$, so $H(X_{t+1})=\log K$, and $\mathcal{L}^\star_t[\mathbf{r}]=H(X_{t+1}\mid X_{\leq t})$ by \autoref{app:in-context}; hence
\begin{align}\label{eq:gain-information}
  \log K-\mathcal{L}^\star_t[\mathbf{r}] &= I(X_{t+1};X_{\leq t}),
  \\
  T\,G(K,T) &= T\log K - H(X_{1:T}) = T\left[\log K-\overline{H(\mathbf{p})}\right] - I(\mathbf{p}_\star;X_{1:T}),
\end{align}
by the chain rule, and $H(X_{1:T}) = H(X_{1:T}\mid\mathbf{p}_\star) + I(\mathbf{p}_\star;X_{1:T})$ with $H(X_{1:T}\mid\mathbf{p}_\star)=T\,\overline{H(\mathbf{p})}$.
The gain per token is thus the entropy gap between the uniform distribution and a random topic, $1-\gamma_{\mathrm{E}}-1/(2K)+O(K^{-2})$ by \autoref{eq:dirichlet-entropy}, minus the information about its own topic that the document leaves unrevealed.
Its exact value follows from \autoref{eq:in-context-exact}, and its two regimes are
\begin{align}\label{eq:gain-regimes}
  G(K,T) &= \log K-\overline{H(\mathbf{p})} - \frac{K-1}{2}\,\frac{\log T}{T} + O_K\!\left(\frac{1}{T}\right) && (T\gg K), \nonumber\\
  G(K,T) &= (2\log 2-1)\,\frac{T-1}{2K} + O\!\left(\frac{T^2}{K^2}\right) && (K\gg T).
\end{align}
The first follows by summing \autoref{eq:in-context-expansion} over positions, which gives $I(\mathbf{p}_\star;X_{1:T})=\frac{K-1}{2}\log T+O_K(1)$: a long context reveals its topic up to the usual parametric cost.
The second follows from \autoref{eq:in-context-exact} at $t\ll K$: a given token appears once in the context with probability $\mathbb{P}(C_a\,{=}\,1)=(K-1)\,t/[(t+K-1)(t+K-2)]\simeq t/K$ and more than once with probability $O(t^2/K^2)$, so that $\mathcal{L}^\star_t[\mathbf{r}]=\log(t+K)-\frac{K}{t+K}\,2\log2\;\mathbb{P}(C_a\,{=}\,1)+O(t^2/K^2)=\log K-(2\log2-1)\,t/K+O(t^2/K^2)$; in a short context most tokens are new, and the gain comes from the few repeated ones.

\section{The spectral baseline}\label{app:spectral}

This appendix supports \autoref{ssec:spectral}: the Hankel formulation of spectral learning, the spectra of the pair and four-token moments under the random topic model, the noise of their empirical estimates, and the resulting sample complexities.
Its asymptotic statements are in the joint limit $D,K\to\infty$ with $D/K$, respectively $D/K^2$, fixed.

\paragraph{Hankel matrix and subspace.}
For the token process, the Hankel matrix has entries $H(u,v):=\mathbb{P}(X_{1:|u|+|v|}=uv)$, indexed by a prefix string $u$ and a suffix string $v$; since tokens are i.i.d. given the topic, $H(u,v)=\frac1D\sum_z g_z(u)\,g_z(v)$ with $g_z(u)=\prod_s p_z(u_s)$, so that $\operatorname{rank}H\leq D$~\citep{balle2014methods}.
Spectral learning first recovers the column space of $H$ from a complete sub-block, i.e. one of rank $D$, then the observation operators from the block with one more symbol; for the random topic model the hidden state is constant within a document, the transition matrix is the identity and the observation matrix is $P$~\citep{hsu2012spectral}.
The block on prefixes and suffixes of $m$ tokens is $M_{2m}=\frac1D\sum_z(\mathbf{p}_z^{\otimes m})(\mathbf{p}_z^{\otimes m})^\top$, whose nonzero spectrum is that of the $D\times D$ Gram matrix $\frac1D[(\mathbf{p}_i^\top\mathbf{p}_j)^m]_{ij}$.
The lifted topics $\mathbf{p}_z^{\otimes m}$ lie in the space of symmetric tensors, of dimension $s_m=\binom{K+m-1}{m}$, and span it as $\mathbf{p}_z$ ranges over the simplex, since a homogeneous polynomial vanishing on the simplex vanishes identically; some $s_m\times s_m$ minor of the lifted topic matrix is therefore a nonzero polynomial in the topic entries, and $\operatorname{rank}M_{2m}=\min\{D,s_m\}$ almost surely for topics drawn from the simplex.
The single-token block $M_2$ is thus complete for $D\leq K$, the two-token block $M_4$ for $D\leq K(K+1)/2$, and so on; the topics follow from $M_3$, $M_5$, \dots, for $D\leq K$ through the whitening of $M_3$ by $M_2$ and the decomposition of the resulting orthogonal tensor~\citep{anandkumar2014tensor}, as recalled in \autoref{app:bayes-learner}.

\begin{figure}[h!]
  \centering
  \includegraphics[width=\linewidth]{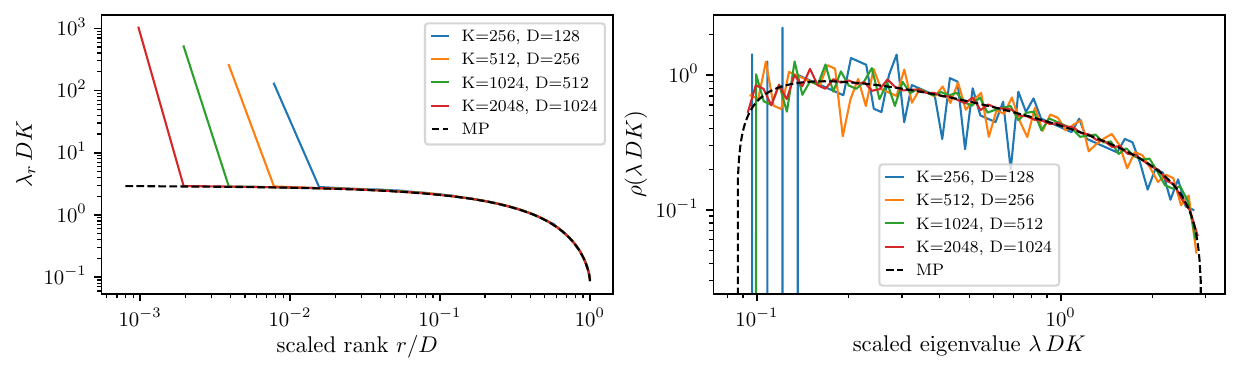}
  \vspace{-10pt}
  \caption{
   \textbf{Population spectrum of $M_2$ for $D\leq K$.}
   \textbf{Left:} ranked eigenvalues $\lambda_r\,DK$ against the scaled rank $r/D$ at $\alpha\,{=}\,D/K\,{=}\,1/2$, for $K\in\{256,512,1024,2048\}$ and one topic realisation per $(K,D)$; dashed: the ranked Marchenko--Pastur law; the spike is the outlier at $\lambda_1 DK\approx D$.
   \textbf{Right:} density of the bulk eigenvalues, spike removed, on the $\lambda\,DK$ scale, against the Marchenko--Pastur density (dashed).
  }
  \label{fig:M2-spectrum}
\end{figure}

\paragraph{Spectrum of $M_2$.}
By \autoref{eq:dirichlet-moments}, $\overline{p_ap_b}=1/(K(K+1))$ for $a\neq b$ and $\overline{p_a^2}=2/(K(K+1))$, so the mean of $M_2$ over the topics is $(I+\mathbf{1}\mathbf{1}^\top)/(K(K+1))$, with one eigenvalue $1/K$ along $\mathbf{1}$ and $1/(K(K+1))$ on its complement.
For a realisation, write $\mathbf{p}_z=\mathbf{1}/K+\boldsymbol{\xi}_z$ with $\boldsymbol{\xi}_z$ centred, orthogonal to $\mathbf{1}$, of covariance $\frac{1}{K(K+1)}(I-\mathbf{1}\mathbf{1}^\top/K)$ and independent across topics; the nonzero spectrum of $M_2=\frac1DPP^\top$ is that of $\frac1DP^\top P=\frac1D\big[\frac1K\mathbf{1}\mathbf{1}^\top+\Xi^\top\Xi\big]$ with $\Xi=[\boldsymbol{\xi}_1,\dots,\boldsymbol{\xi}_D]$.
In the $(K-1)$-dimensional complement of $\mathbf{1}$, the rescaled columns $\sqrt{K(K+1)}\,\boldsymbol{\xi}_z$ are independent, isotropic and log-concave, being uniform on a convex set, so that as $D,K\to\infty$ with $D/(K-1)\to\alpha\in(0,1)$ the eigenvalues of $\frac{K(K+1)}{K-1}\Xi^\top\Xi$ follow the Marchenko--Pastur law of ratio $\alpha$, with extreme eigenvalues converging to the edges $(1\pm\sqrt\alpha)^2$~\citep{pajor2009limiting,chafai2018extremal}.
The rank-one term removes at most one eigenvalue from this bulk, by interlacing, and produces the outlier $D/K$; dividing by $D$,
\begin{align}\label{eq:M2-spectrum}
  \lambda_1(M_2) = \frac{1}{K}\,(1+o_{\mathbb{P}}(1)),
  \qquad
  \lambda_{2}(M_2),\ \lambda_D(M_2) = \frac{K-1}{DK(K+1)}\Big(1\pm\sqrt{\tfrac{D}{K-1}}\Big)^2(1+o_{\mathbb{P}}(1)),
\end{align}
and the remaining $K-D$ eigenvalues vanish: one spike at $1/K$ and a bulk of $D-1$ eigenvalues at scale $1/(DK)$, as in \autoref{fig:M2-spectrum}.

\begin{figure}[h!]
  \centering
  \includegraphics[width=\linewidth]{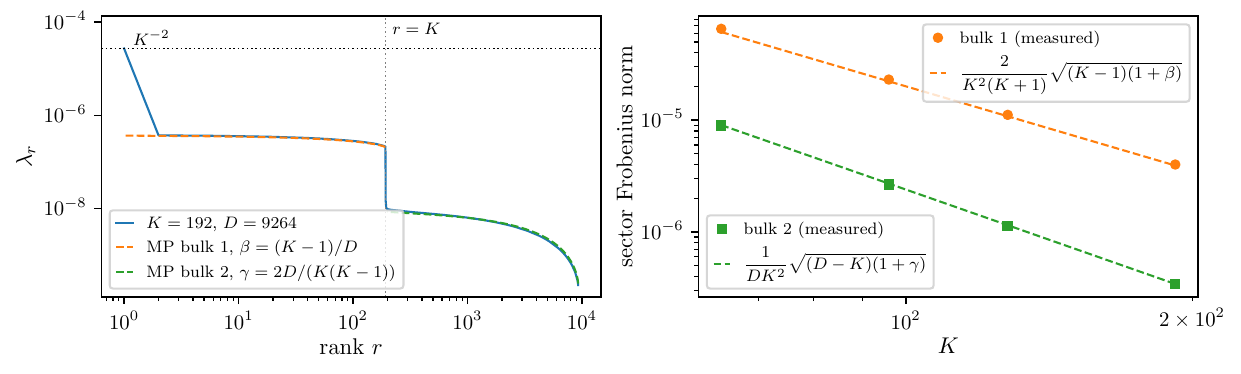}
  \vspace{-10pt}
  \caption{
   \textbf{Population spectrum of $M_4$ for $K<D\leq K(K+1)/2$.}
   \textbf{Left:} ranked eigenvalues of $M_4$ for one realisation at $K\,{=}\,192$, $D\,{=}\,K(K+1)/4\,{=}\,9264$: the spike at $K^{-2}$ (dotted), the $K-1$ eigenvalues of the linear block and, beyond $r\,{=}\,K$ (dotted), the $D-K$ eigenvalues of the quadratic block, against the Marchenko--Pastur predictions of the two blocks (dashed).
   \textbf{Right:} Frobenius norm of each block against $K$, for $D\,{=}\,K(K+1)/4$, with the parameter-free predictions (dashed).
  }
  \label{fig:M4-spectrum}
\end{figure}

\paragraph{Spectrum of $M_4$.}
For $K<D\leq K(K+1)/2$ the complete block is $M_4$, whose nonzero spectrum is that of $\frac1D[(\mathbf{p}_i^\top\mathbf{p}_j)^2]_{ij}$.
Writing $\mathbf{p}^{\otimes2}=\mathbf{u}^{\otimes2}+(\mathbf{u}\otimes\boldsymbol{\xi}+\boldsymbol{\xi}\otimes\mathbf{u})+\boldsymbol{\xi}^{\otimes2}$ with $\mathbf{u}=\mathbf{1}/K$ separates three scales.
The mean feature has squared norm $(K+3)/(K(K+1)^2)\simeq K^{-2}$ and gives one spike at $K^{-2}$; the linear features span $K-1$ directions with covariance eigenvalue $2/(K^2(K+1))$; the centred quadratic features span the remaining $r_2=K(K-1)/2$ directions, with average covariance eigenvalue $2K^{-4}(1+O(K^{-1}))$.
Neglecting the coupling between the blocks, a Marchenko--Pastur approximation of each predicts $K-1$ eigenvalues of order $K^{-3}$, with edges $\frac{2}{K^2(K+1)}(1\pm\sqrt{(K-1)/D})^2$, and $D-K$ eigenvalues of order $1/(DK^2)$, with edges $\frac{1+O(K^{-1})}{DK^2}(1\pm\sqrt{D/r_2})^2$ when $D/r_2\to\beta\in(0,1)$.
This block picture is heuristic, since the quadratic features are neither isotropic nor log-concave and the coupling is neglected; \autoref{fig:M4-spectrum} compares it with the spectrum of one realisation at $D=K(K+1)/4$, where both bulks follow the predicted curves and their Frobenius norms agree with the predictions within $7\%$ for the linear block and $2\%$ for the quadratic one, for $K$ from $64$ to $192$.

\begin{figure}[h!]
  \centering
  \includegraphics[width=\linewidth]{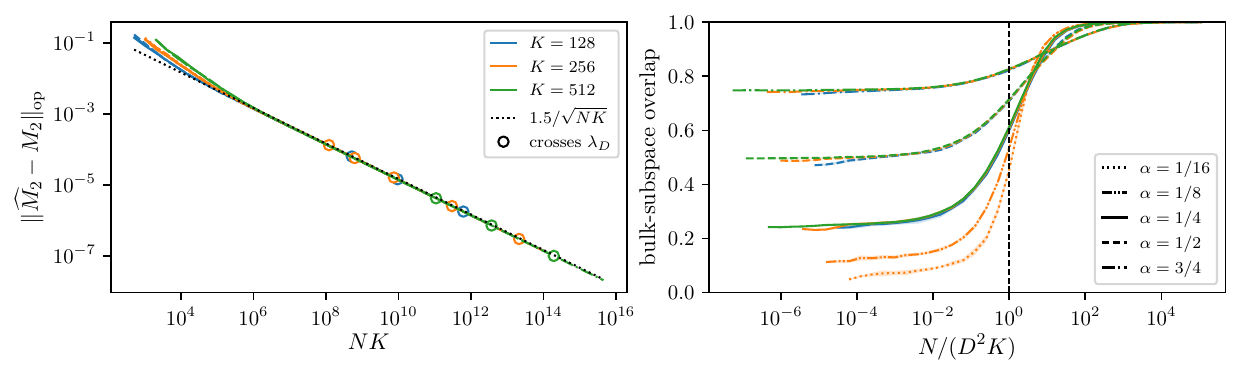}
  \vspace{-10pt}
  \caption{
   \textbf{Convergence of the empirical pair moment for $D\leq K$.}
   \textbf{Left:} operator-norm error $\|\widehat M_2-M_2\|_{\mathrm{op}}$, geometric mean over eight replicas, against $NK$ for eleven $(K,D)$ configurations, $\alpha\,{=}\,D/K\in\{1/4,1/2,3/4\}$ times $K\in\{128,256,512\}$ plus $K\,{=}\,256$ with $D\in\{16,32,64,128,192\}$; colour encodes $K$, line style $\alpha$; all curves collapse on the dotted $1.5/\sqrt{NK}$; circles mark where each curve crosses its own population $\lambda_D$, beyond which the lower bulk edge is resolved.
   \textbf{Right:} overlap $\frac{1}{D-1}\|U_{2:D}^\top\widehat U_{2:D}\|_F^2$ between the bulk eigenvector subspaces of $M_2$ and $\widehat M_2$, mean $\pm$ one standard deviation over replicas, against $N/(D^2K)$; each curve rises from the random-subspace value $(D-1)/(K-1)$ towards one around $N/(D^2K)=O(1)$ (dashed line at one).
   One topic realisation per $(K,D)$, uniform topic prior.
  }
  \label{fig:M2-convergence}
\end{figure}

\paragraph{Noise and sample complexity.}
Let $\widehat M_2$ be the empirical pair moment from $N$ independent pairs, so that $\widehat M_2-M_2=\frac1N\sum_{\mu}Y_\mu$ with $Y_\mu=\mathbf{x}_1^{(\mu)}\mathbf{x}_2^{(\mu)\top}-M_2$ centred, $\|Y_\mu\|_{\mathrm{op}}\leq1+1/K$ and $\mathbb{E}[Y_\mu Y_\mu^\top]\preceq\operatorname{diag}(\bar{\mathbf{p}})$, with $\bar{\mathbf{p}}=\frac1D\sum_z\mathbf{p}_z$ the unigram marginal, of entries close to $1/K$.
The matrix Bernstein inequality~\citep{tropp2012userfriendly} then gives, with probability at least $1-\delta$,
\begin{align}\label{eq:M2-bernstein}
  \|\widehat M_2-M_2\|_{\mathrm{op}} \leq \sqrt{\frac{2\log(2K/\delta)}{KN}}+\frac{2\log(2K/\delta)}{3N},
\end{align}
a noise level of order $(NK)^{-1/2}$; the measured error follows $1.5/\sqrt{NK}$ across eleven $(K,D)$ configurations (\autoref{fig:M2-convergence}, left).
Treating the noise as a generic perturbation, an eigenvalue $\lambda$ of $M_2$ is resolved once $\lambda\gtrsim1.5/\sqrt{NK}$: the spike at $N\sim K$, and the lower edge of the bulk, $\frac{K-1}{DK(K+1)}(1-\sqrt\alpha)^2$, at $N^*=2.25\,D^2K/(1-\sqrt\alpha)^4$, i.e. at $N\sim D^2K$ up to an $\alpha$-dependent constant.
This criterion is heuristic, but the scale is confirmed by the eigenvectors: the overlap between the bulk eigenvector subspaces of $\widehat M_2$ and $M_2$ rises from its random value $(D-1)/(K-1)$ to one around $N/(D^2K)=O(1)$ (\autoref{fig:M2-convergence}, right).
For the four-token moment the same bound has a variance proxy of order $K^{-2}$ and gives a noise level of order $(NK^2)^{-1/2}$, so the spike at $K^{-2}$ is resolved at $N\sim K^2$, the linear block at $N\sim K^4$ and the quadratic block, of scale $1/(DK^2)$, at $N\sim D^2K^2$; extrapolating, the bulk of $M_{2m}$ sits at $1/(DK^m)$ against a noise of $(NK^m)^{-1/2}$, and its resolution takes $N\sim D^2K^m$.

\section{Experimental details}\label{app:experiments}

This appendix specifies the transformer experiments of \autoref{sec:experiments}.
The code to reproduce them is available at \url{https://github.com/fracagnetta/random-topic-model}.

\paragraph{Data.}
Every run draws its own topic matrix $P$, with $D$ topics from $\mathrm{Dir}_K(\mathbf{1})$ and uniform topic weights, and from it $N$ training documents and $32\,768$ validation documents; a further $32\,768$ OOD documents are generated from $16$ fresh topics, drawn independently of $P$ from the same prior.
Each point of a learning curve averages four runs with independent topic matrices, test ID and OOD corpora and initialisation, and error bars are standard errors over the four.
Documents carry $T{+}1$ tokens: the model reads the first $T$ and is trained to predict tokens $2$ to $T{+}1$, so that its loss profile covers context lengths $1$ to $T$, the $T$-gram loss is the loss at context length $T$, and the autoregressive loss is the average of the profile over the $T$ positions.

\paragraph{Model.}
A one-block decoder-only transformer: token embeddings of dimension $d_{\rm emb}$ with no positional embedding, so that the causal mask is the only source of order, as befits documents that are exchangeable given their topic; one causal self-attention head; a ReLU multilayer perceptron with one hidden layer of width $4d_{\rm emb}$; both sub-layers pre-normalised and residual; a final layer normalisation and an untied linear readout to the $K$ logits.
Attention and perceptron carry no biases and there is no dropout.
The model has $12\,d_{\rm emb}^2+2Kd_{\rm emb}+O(d_{\rm emb})$ parameters, from $5\times10^4$ at $(K,d_{\rm emb})\,{=}\,(32,64)$ to $3\times10^6$ at $(128,512)$.
Weights are initialised as standard normals and each linear map is scaled by the inverse square root of its input dimension, the readout by $1/d_{\rm emb}$.

\paragraph{Training.}
The loss is the cross-entropy of the next token, averaged over the $T$ positions and the documents of a batch.
Batches of $b$ documents are taken cyclically from the corpus and the parameters are updated with Adam, moment parameters $(0.9,0.999)$ and no weight decay, at a learning rate $\eta$ raised linearly from zero over the first $8$ steps and constant thereafter, for a fixed number of passes over the corpus.

\begin{figure}[h!]
  \centering
  \includegraphics[width=\linewidth]{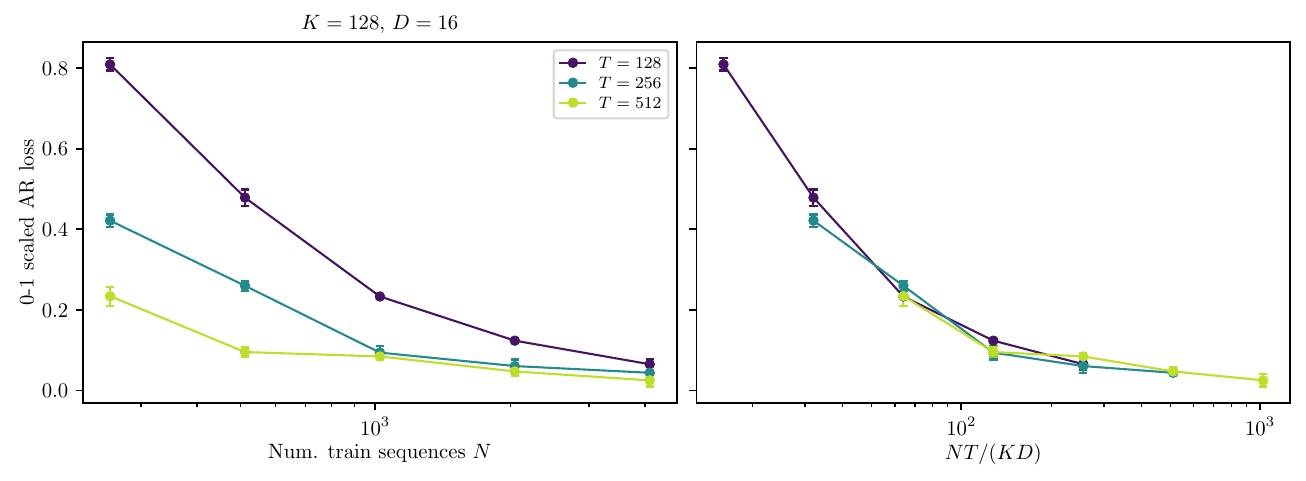}
  \vspace{-10pt}
  \caption{
   \textbf{Learning curves at different document lengths collapse on the number of tokens per parameter.}
   \textbf{Left:} Scaled ID autoregressive loss $(\mathcal{L}_{\rm AR}(N)-\mathcal{L}_{\rm AR}(\infty))/(\log{K}-\mathcal{L}_{\rm AR}(\infty))$ versus number of training sequences for one-block transformers trained on random topic model data with $K\,{=}\,128$, $D\,{=}\,16$ and $T\,{=}\,128,256,512$ (colour).
   $\mathcal{L}_{\rm AR}(\infty)$ is the autoregressive loss of the specialised predictor at each $T$, averaged over $32$ realisations of the topic matrix.
   \textbf{Right:} Against $NT/(DK)$ the three curves coincide: at fixed $N$, longer documents bring the loss closer to its floor in proportion to the number of tokens they contribute.
  }
  \label{fig:Tscaling}
\end{figure}

\begin{figure}[h!]
  \centering
  \includegraphics[width=\linewidth]{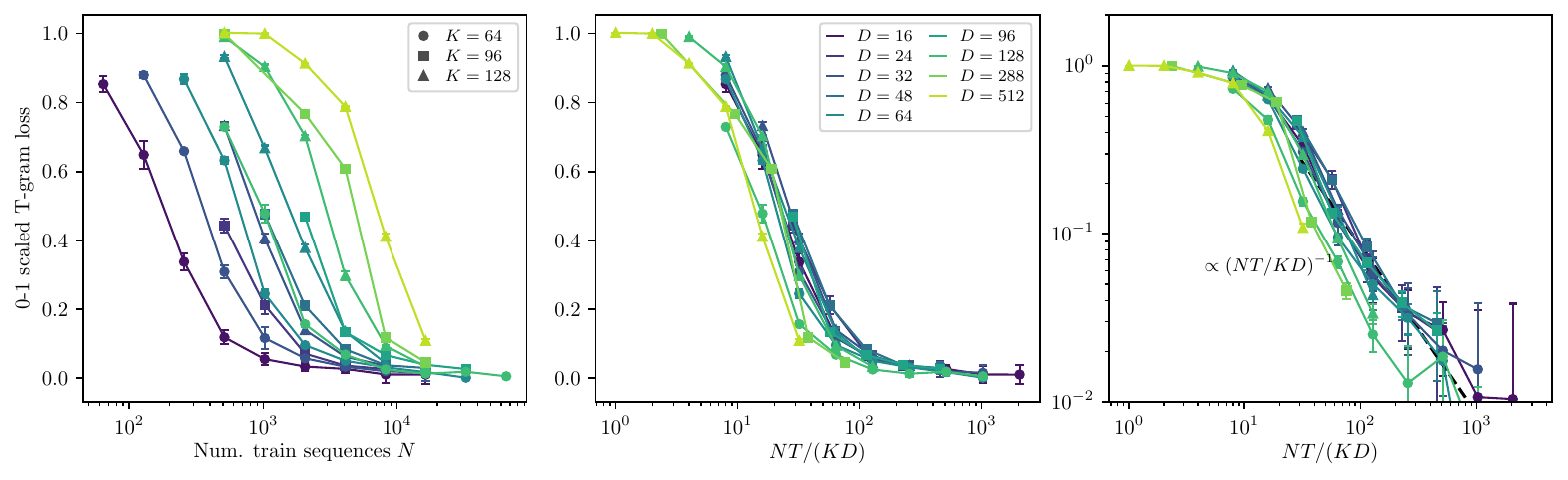}
  \vspace{-10pt}
  \caption{
   \textbf{Transformers match the Bayes-optimal specialisation learning curve at the longest context too.}
   $T$-gram counterpart of \autoref{fig:learning-curve-scaling}, for the same twelve $(K,D)$ at $T\,{=}\,128$.
   \textbf{Left:} Scaled ID $T$-gram loss $(\mathcal{L}_{T}(N)-\mathcal{L}_{T}(\infty))/(\log{K}-\mathcal{L}_{T}(\infty))$ versus number of training sequences, where $\mathcal{L}_{T}(\infty)$ is the mean entropy $\frac1D\sum_z H(\mathbf{p}_z)$ of the topics of each run, the large-context floor of the specialised predictor.
   \textbf{Middle:} Rescaling the $x$-axis as the number of tokens per parameter of the data-generating process collapses all the curves onto a single functional form, more tightly than for the autoregressive loss.
   \textbf{Right:} The decay towards $0$ (black dashed) follows the $x^{-1}$ rate of the Bayes-optimal learning curve.
  }
  \label{fig:tgram-scaling}
\end{figure}

\paragraph{Evaluation and early stopping.}
The validation autoregressive loss is measured at log-spaced steps throughout training.
The checkpoint of lowest validation loss defines early stopping, and its loss is the value that enters the ID learning curves (\autoref{fig:main}, \autoref{fig:learning-curve-scaling}, \autoref{fig:Tscaling}).
At a coarser log-spaced subset of the steps the full loss profiles on the validation and on the OOD documents are recorded as well.
The $T$-gram ID loss of a run (\autoref{fig:tgram-scaling}) is the minimum over these recorded checkpoints of the loss at context length $T$; the OOD losses are read either at the recorded checkpoint of lowest validation loss, ``at early stopping'' (\autoref{fig:main}, \autoref{fig:ood-window}), or minimised over the recorded checkpoints, ``best over training'' (\autoref{fig:main}, \autoref{fig:ood-scaling}), as stated in each caption.

\paragraph{Hyperparameter selection.}
The embedding dimension, batch size, learning rate and training budget are selected separately for every $(K,D,T,N)$ on the validation loss at early stopping, so that each learning curve reports the best model of the family rather than a fixed architecture, and the model is not the bottleneck of the comparison with the Bayes-optimal learning curves.
The grid is $d_{\rm emb}$ from $K$ to $512$ in powers of two, $b\in\{16,32,64,128\}$ and $\eta\in\{10^{-2},3\times10^{-2},10^{-1}\}$; larger dictionaries require larger $d_{\rm emb}$, and the selected width grows with $D$ at fixed $K$.
The number of passes is extended until the early-stopping checkpoint falls well inside the budget, or until a doubling of the budget improves the validation loss by less than the spread over seeds.
A candidate combination replaces the incumbent only when its early-stopping validation loss, over four fresh seeds, is lower beyond the standard errors, or tied with a clearly earlier early-stopping step.
The combination selected on the ID validation loss serves for every quantity reported at that $(K,D,T,N)$, including the OOD losses, which play no role in the selection.

\section{Additional figures}\label{app:extra-figures}\label{app:varying-T}

\begin{figure}[h!]
  \centering
  \includegraphics[width=\linewidth]{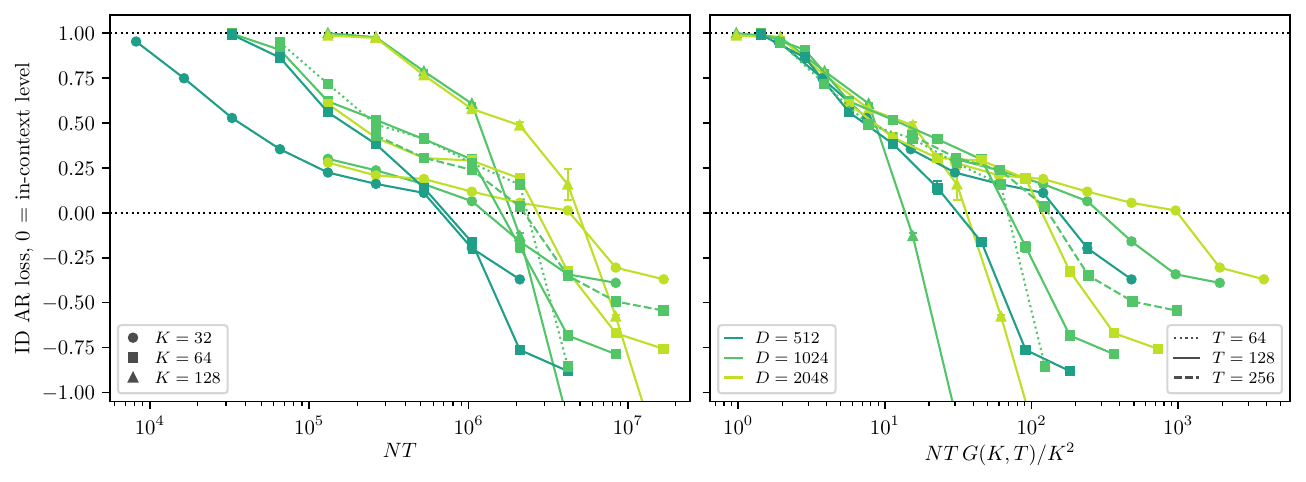}
  \vspace{-10pt}
  \caption{
   \textbf{The ID loss descends to the in-context level at the predicted sample scale.}
   \textbf{Left:} ID autoregressive loss at early stopping, scaled as $(\mathcal{L}_{\rm AR}(N)-\mathcal{L}^\star_{\rm AR}[\mathbf{r}])/(\log{K}-\mathcal{L}^\star_{\rm AR}[\mathbf{r}])$, where $\mathcal{L}^\star_{\rm AR}[\mathbf{r}]\,{=}\,\log K-G(K,T)$ is the autoregressive loss of the in-context predictor, the same on ID and OOD documents (\autoref{app:in-context}).
   This scaled loss is $1$ at the uniform prediction, $0$ at the in-context solution, and negative at specialisation.
   Versus number of training tokens $NT$, for the $(K,D)$ and $T$ of \autoref{fig:ood-scaling}, with the same line styles.
   \textbf{Right:} When plotted against $NT\,G(K,T)/K^2\,{=}\,N/(2N_{\rm icl})$ (\autoref{eq:n-icl}) the descents to the in-context level coincide for all $K$, $D$ and $T$, with no adjustable constant.
   The subsequent departure below zero is specialisation.
  }
  \label{fig:id-kscaling}
\end{figure}

\end{document}